\documentclass[remotesensing,article,accept,moreauthor,10pt,a4paper]{mdpi} 
\usepackage{longtable, xcolor}
\usepackage{epstopdf}
\firstpage{1} 
\pubvolume{11}
\issuenum{1}
\articlenumber{5}
\pubyear{2019}
\copyrightyear{2019}
\doinum{10.3390/rs11050493 }
\history{Received: 27 December 2018; Accepted: 24 February 2019; Published: 28 February 2019}

\Title{Aggregated Deep Local Features for Remote Sensing Image Retrieval}

\Author{Raffaele Imbriaco $^{1,\ddagger}$*, Clint Sebastian $^{1,\ddagger}$, Egor Bondarev $^{1}$ and Peter H.N. de With $^{1}$}

\AuthorNames{Raffaele Imbriaco, Clint Sebastian, Egor Bondarev and Peter H.N. de With}

\address{%
$^{1}$ \quad Video Coding and Architectures Group, Department of Electrical Engineering, Eindhoven University of Technology; Eindhoven 5612 AZ; c.sebastian@tue.nl, e.bondarev@tue.nl, p.h.n.de.with@tue.nl\\
}

\corres{Correspondence: r.imbriaco@tue.nl}

\secondnote{These authors contributed equally to this work.}

\abstract{Remote Sensing Image Retrieval remains a challenging topic due to the special nature of Remote Sensing imagery. Such images contain various different semantic objects, which clearly complicates the retrieval task. In this paper, we present an image retrieval pipeline that uses attentive, local convolutional features and aggregates them using the Vector of Locally Aggregated Descriptors (VLAD) to produce a global descriptor. We study various system parameters such as the multiplicative and additive attention mechanisms and descriptor dimensionality. We propose a query expansion method that requires no external inputs. Experiments demonstrate that even without training, the local convolutional features and global representation outperform other systems. After system tuning, we can achieve state-of-the-art or competitive results. Furthermore, we observe that our query expansion method increases overall system performance by about 3\%, using only the top-three retrieved images. Finally, we show how dimensionality reduction produces compact descriptors with increased retrieval performance and fast retrieval computation times, e.g. 50\% faster than the current systems.
}


\begin{document}
\section{Introduction}

Content Based Image Retrieval (CBIR) is a branch of image retrieval that identifies similar or matching images in a database using visual query information (e.g. color, textures). This technique has been improving steadily, evolving from using hand-engineered descriptors, such as SIFT~\cite{lowe1999object}, to learned convolutional descriptors extracted from Convolutional Neural Networks (CNNs)~\cite{babenko2014neural}\cite{sunderhauf2015place}\cite{noh2017large}. Furthermore, several descriptor aggregation methods have been proposed for large-scale image retrieval such as Bag-of-Words (BoW)~\cite{sivic2003video}, Vector of Locally Aggregated Descriptors (VLAD)~\cite{jegou2010aggregating}, Fisher Kernels (FK)~\cite{perronnin2007fisher} and Memory Vectors~\cite{iscen2017memory}. In the area of Remote Sensing (RS), similar developments have occurred and associated techniques have been employed. Increasing spatial resolution and image database size have fueled interest in systems and techniques capable of efficiently retrieving RS images based on their visual content. Methods for Remote Sensing Image Retrieval (RSIR) have incorporated some of the advances of standard image retrieval, with methods using the BoW representation~\cite{bai2014bag}, locally invariant descriptors~\cite{yang2013geographic} or learned convolutional features~\cite{penatti2015deep}\cite{tang2018unsupervised}. 

\begin{figure}[H]
\centering
\includegraphics[width=\textwidth]{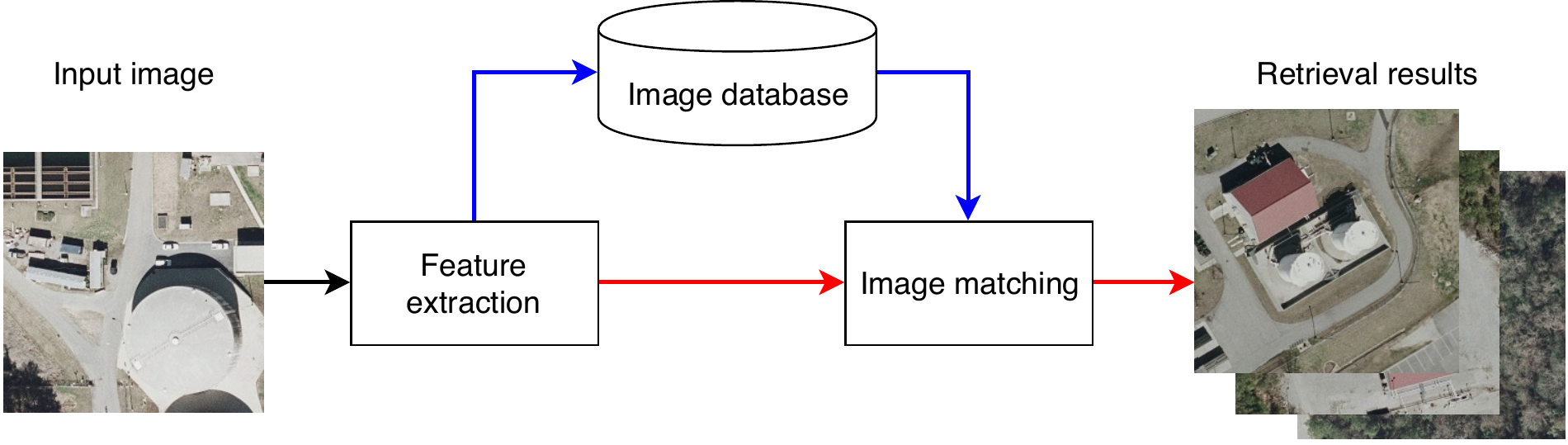}
\caption{Flowchart of a CBIR system showing query (red) and database (blue) dataflows.}
\label{fig:flowchart}
\end{figure}

A CBIR system can be outlined with two basic processes. The first, feature extraction, deals with the generation of compact, descriptive representations of the visual content of the image. This process maps the images into a high-dimensional feature space. The second, image matching, produces a numerical value that represents the similarity of the visual content of the query and database images. This value is used to sort the images and select the best candidate matches, which are then presented to the user. This generic CBIR system is depicted in Figure~\ref{fig:flowchart}. 
  
The challenge of RSIR is that RS images depict large geographic areas, which may contain a wide range of semantic objects. Furthermore, images belonging to the same semantic category may have significant appearance variations. Thus, accurately retrieving images with similar visual content requires extraction of sufficiently descriptive and robust features. Additionally, the system should provide its output within a reasonable time frame. Thereby, descriptor generation should produce highly discriminative representations, while simultaneously enabling efficient database search. Literature divides descriptors into three categories. The first category consists of low-level descriptors that encode information like color, spectral features and texture~\cite{manjunath1996texture,risojevic2013fusion,shyu2007geoiris}. The second category is represented by mid-level features which are generally obtained by means of local, invariant features aggregated into global representations using BoW as in~\cite{yang2013geographic}. The third category encompasses semantic features that encode semantic concepts like ``tree'' or ``building'' and which are extracted from CNNs. Recently, the latter type has become popular in the field of image retrieval and has been successfully applied for landmark retrieval~\cite{babenko2014neural}\cite{babenko2015aggregating}\cite{hou2017evaluation}, visual place recognition~\cite{sunderhauf2015place}\cite{arandjelovic2016netvlad} and RSIR~\cite{bai2014bag}\cite{tang2018unsupervised}. In RSIR, we did not find any systems that directly generate local convolutional descriptors for the most salient or representative regions of images. 

In our work, we present a novel RSIR system that, to our knowledge, is the first to use attentive, multi-scale convolutional features. Unlike the previous work, the local descriptors are learned using attention mechanisms which do not require partitioning the image into small patches. The attention mechanisms are discussed later. This enables us to extract descriptors for the most relevant objects of an image, while retaining the ability to capture features at various scales. Additionally, we perform descriptor aggregation with Vector of Locally Aggregated Descriptors (VLAD), instead of using the BoW framework which is regularly applied in other RSIR methods. Aggregating local descriptors using VLAD results in better retrieval performance~\cite{jegou2010aggregating}. Our system also utilizes query expansion to improve retrieval accuracy. This paper provides the following contributions.

\begin{itemize}
    \item A novel RSIR pipeline that extracts attentive, multi-scale convolutional descriptors, leading to state-of-the-art or competitive results on various RSIR datasets. These descriptors automatically capture salient, semantic features at various scales. 
    \item We demonstrate the generalization power of our system and show how features learned for landmark retrieval can successfully be re-used for RSIR and result in competitive performance.
    \item We experiment with different attention mechanisms for feature extraction and evaluate whether context of the local convolutional features is useful for the RSIR task.
    \item We introduce a query expansion method that requires no user input. This method provides an average performance increase of approximately 3\%, with low complexity. 
    \item RSIR with lower dimensionality descriptors is evaluated. Reduction of the descriptor size yields better retrieval performance and results in a significant reduction of the matching time.
\end{itemize}

The paper is organized as follows. Section~\ref{s2} presents and analyzes the related work both in  standard image retrieval and RSIR. Section~\ref{s3} describes the architectural pipeline of the presented RSIR system. This includes the CNN architecture and its training, descriptor extraction, descriptor aggregation, and image retrieval. Section~\ref{s4} introduces the datasets, evaluation metrics, and experimental results. Finally,  Section~\ref{s5} concludes the paper. 

\section{Related Work}
\label{s2}
This section summarizes various works both in standard image retrieval and RSIR. First, an overview of the trends in traditional image retrieval is given, particularly emphasizing on the descriptor aggregation techniques for large-scale CBIR and extraction of convolutional descriptors. Afterwards, various RSIR methods are presented, ranging from the traditional techniques (i.e. hand-engineered features) to the current state-of-the-art methods.
\subsection{Image retrieval}
\subsubsection{Descriptor aggregation}
Until recent years, image retrieval has been based on handcrafted features, such as  SIFT~\cite{lowe1999object}, which are commonly combined with aggregation methods, e.g. Bag-of-Words (BoW)~\cite{sivic2003video}. Whereas local descriptors present some desirable traits, such as being invariant to transformations and enabling geometric verification. However, they lack the compactness and memory efficiency provided by global descriptors. Aggregation is necessary for efficient database search, since a single image can contain a large number of local descriptors. Descriptor aggregation techniques, like BoW, produce a single global representation per image. This representation can be, for example, a frequency  histogram of visual words in an image or a high-dimensional vector. The similarity between the query and database images is determined by a vector operation, like the Euclidean distance or cosine similarity. Given a large number of descriptors in an image, the objective of an aggregation method is to generate a compact, yet descriptive global representation. Other descriptor aggregation methods have been proposed in literature. Among these, we consider VLAD~\cite{jegou2010aggregating} and Fisher Kernels (FKs)~\cite{perronnin2007fisher} as the principal techniques. A common factor across these descriptor aggregation methods is that they first compute a quantizer or codebook. For an FK codebook, this is implemented by means of a Gaussian Mixture Model, while for VLAD and BoW, the k-means algorithm is used. In the case of VLAD and FK, the statistics of the local descriptors are encoded into a single vector, yielding better performance than the BoW model~\cite{jegou2010aggregating}. 

Traditional image retrieval utilizes post-processing techniques which increase retrieval performance. Geometric verification~\cite{sivic2003video}\cite{jegou2008hamming, lowe2004distinctive, philbin2007object} and query expansion\cite{Voorhees:1994:QEU:188490.188508, chum2007total, chum2011total, arandjelovic2012three} are two popular methods in literature for such enhancement. The first technique, called geometric verification, identifies features that appear both in the query and the candidate images. The RANSAC\cite{fischler1981random} or Hamming Embedding \cite{jegou2008hamming}\cite{jain2011asymmetric} algorithms are typically deployed, where the quality of the candidates is determined by the amount of matching features. This enables the retrieval system to reject candidates with similar visual content but indicating to different objects. The second technique, called query expansion, uses the features of the correct matches to expand the features of the query to support successful matching. In this manner, more relevant images can be retrieved with successive queries. Correctness of a candidate image is usually assessed with geometric verification. The query expansion technique is particularly useful in combination with geometric verification, while the latter one can be used independently. 

In the work discussed above, several local descriptors are generated per image, requiring aggregation for efficient retrieval. A different solution is to directly extract global representations from the images, e.g. GIST descriptors~\cite{oliva2001modeling}. However, there is a noticeable trend in image retrieval literature in which handcrafted descriptors are being replaced by high-level learned descriptors extracted from CNNs. These high-level descriptors provide discriminative, global representations that outperform handcrafted descriptors~\cite{radenovic2018revisiting}. 

\subsubsection{Convolutional descriptors}
In~\cite{babenko2014neural}, Babenko \emph{et al.} demonstrate that the high-level features that CNN systems employ for classification can also be used for image retrieval. The retrieval performance of such a system can be increased by retraining the CNN with images relevant to the retrieval task. They also show that features extracted from mid-level layers are better for image retrieval than those extracted from deeper layers. However, convolutional layers produce high-dimensional tensors that require aggregation into more compact representations for efficient image retrieval. The most common approach is to vectorize the feature tensor using pooling operations, like sum-pooling~\cite{babenko2015aggregating} or max-pooling~\cite{azizpour2015generic}. Tolias~\emph{et al.} in~\cite{tolias2015particular} present an improved sum-pooling operation that uses integral images for efficient descriptor calculation. Their descriptors encode regions in the image and are used for both image retrieval and object localization. More recently in~\cite{noh2017large}, a new CNN architecture has been presented that learns and extracts multi-scale, local descriptors in salient regions of an image. These descriptors, called Deep Local Features (DELF), have been shown to outperform other global representations and handcrafted descriptors for image retrieval~\cite{radenovic2018revisiting}. An advantage presented by DELF descriptors over traditional handcrafted descriptors is that the relevance of a particular feature in the image is also quantified. This is done via an attention mechanism that learns which features are more discriminative for a specific semantic class. 

RSIR can be considered a specialized branch of image retrieval. The objective remains to identify similar images based on their visual content. However, RS imagery presents unique challenges that impede the direct deployment of some techniques common in image retrieval (e.g. geometric verification, query expansion). RS images cover large geographic areas that may contain a diverse amount of different semantic objects, captured at various scales. Additionally, common RS datasets include images belonging to the same semantic class, but they show large appearance variations, or were acquired in different geographic regions. This has resulted in the development of specialized solutions to the RSIR problem, which are presented below.

\subsection{Remote Sensing Image Retrieval}
The earliest RSIR work~\cite{manjunath1996texture}\cite{haralick1973textural} present handcrafted textural features for image matching and classification. Following work from~\cite{datcu1998spatial} and~\cite{schroder1998spatial} presents a more complex system. The authors deploy a Bayesian framework that uses Gibbs-Markov random fields as spatial information descriptors. Bao~\cite{bao2004comparative} studies the effectiveness of various distance metrics (e.g. Euclidean distance, cosine similarity) for histogram-based image representations. A complete CBIR system for RS imagery, GeoIRIS, is presented in \cite{shyu2007geoiris}. This system extracts, processes, and indexes a variety of image features. It uses patch-based features (spectral, texture and anthropogenic) and object-based features (shape, spatial relationships) to provide information-rich representations of RS images. The method proposed in~\cite{ferecatu2007interactive} involves the user by employing Relevance Feedback to train a Support Vector Machine. An improved Active Learning method using Relevance Feedback is presented in \cite{demir2015novel}, where it is attempted to minimize the required number of user annotations necessary. 
In~\cite{yang2013geographic} and~\cite{sun2012automatic}, traditional image retrieval features and frameworks appear for CBIR and target detection. These works use SIFT~\cite{lowe1999object}, Harris~\cite{harris1988combined} or Difference-of-Gaussians~\cite{vidal2003object} features together with the BoW framework to provide compact global representations of images. An alternative aggregated descriptor is introduced in~\cite{risojevic2013fusion}, where SIFT (local) and textural (global) features are fused into a single descriptor and fed to a classifier. More recent work~\cite{demir2016hashing} produces extremely small binary global representations for fast retrieval.  The work discussed above, uses low-level features (color, texture) or mid-level features (BoW histogram) for RSIR. However, these types of features do not encode semantic information. A way to reduce this semantic gap is presented in~\cite{wang2013remote}. This work uses low-level visual features to create more complex semantic interpretations of objects using the Conceptual Neighbourhood Model~\cite{bruns1996similarity}. It constructs a relational graph between scene objects to retrieve images based on semantic information. 

More recent methods have switched from hand-engineered features to convolutional descriptors\cite{marmanis2016deep}. Penatti~\emph{et al.}~\cite{penatti2015deep} demonstrate  that convolutional descriptors generalize well for aerial image classification, even when the descriptors are trained on other datasets. An RSIR system, presented in~\cite{bai2014bag}, learns the mapping from images into their textual BoW representation. Another work that uses convolutional features for RSIR is~\cite{tang2018unsupervised}. It achieves state-of-the-art results by dividing the image into patches and extracting a convolutional descriptor per patch. These descriptors are aggregated using the BoW framework. A key difference between the RSIR system in~\cite{tang2018unsupervised} and other methods using convolutional descriptors is that their training is unsupervised. It deploys a deep auto-encoder architecture that reconstructs the input image patches and learns relevant features. \textcolor{black}{Recent work concentrates on producing compact and discriminative descriptors. This is done either by explicit learning of non-binary short codes as in \cite{Zhou2017LearningLD} or by joint learning of the features and binary representations as in \cite{Li2018LearningSD, Roy2018DeepMA, Li2018LargeScaleRS}. However, these methods produce low-dimensional global descriptors, meaning that irrelevant background information may still be encoded in the final descriptors. Our attention-based feature extraction module can learn what image regions are more relevant to specific classes, disregarding irrelevant regions. Furthermore, our method allows for the generation of lower dimensional vectors using well-known techniques such as Principal Component Analysis (PCA)~\cite{abdi2010principal}. Moderate dimensionality reduction has a positive effect on image retrieval peformance in \cite{jegou2012aggregating}. An inherent advantage of the approach presented in our work is, that the final dimensionality of the descriptors can be selected depending on application-specific requirements without changing the CNN architecture.}

As can be noticed, the advances in RSIR follow from developments in traditional image retrieval. However, many techniques require adaptation to the characteristics of RS imagery. This explains why in this work, we present an RSIR system that combines state-of-the-art local descriptors and techniques for image retrieval. These local convolutional descriptors are extracted in an end-to-end manner based on feature saliency. These are combined with VLAD aggregation to achieve compact and descriptive image representations, which enables fast database search. This approach enables our system to capture high-level semantic information at various scales. 

\section{Architectural pipeline}
\label{s3}
\begin{figure}[H]
\centering
\includegraphics[width=\textwidth]{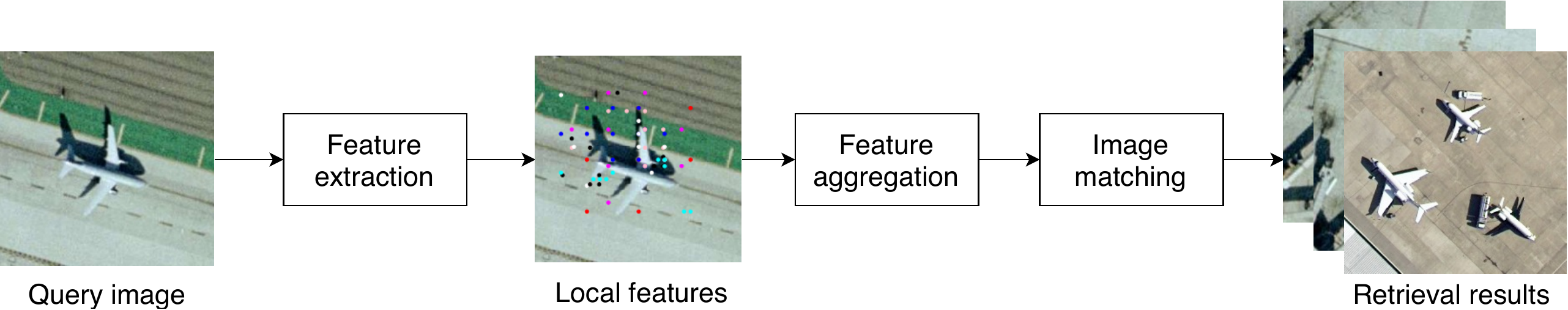}
\caption{Overview of the proposed RSIR pipeline.}
\label{fig:sys_arch}
\end{figure}

This section describes the processes that compose our RSIR system, including feature extraction, descriptor aggregation, and image matching. First, it introduces the Deep Local Descriptor (DELF)~\cite{noh2017large} architecture, the different attention modules utilized in this work, and the training procedure. Second, the formulation of the VLAD representation is summarized. The third part presents the image matching and retrieval process. Additionally, a query expansion strategy is introduced that uses only the global representations and requires no user input. The simplified system architecture is depicted in Figure~\ref{fig:sys_arch}.

\subsection{Feature extraction}
The first step in any CBIR system is to find meaningful and descriptive representations of the visual content of images. Our system makes use of DELF, which is based on local, attentive convolutional features that achieve state-of-the-art results in difficult landmark recognition tasks~\cite{noh2017large}\cite{radenovic2018revisiting}. The feature extraction process can be divided into two steps. First, during feature  generation, dense features are extracted from images by means of a fully convolutional network trained for classification. We use the ResNet50 model~\cite{He_2016_CVPR} and extract the features from the third convolutional block. The second step involves the selection of attention-based key points. The dense features are not used directly but instead, an attention mechanism is exploited to select the most relevant features in the image. Given the nature of RS imagery, irrelevant objects commonly occur in the images which are belonging to different classes. For example, an image depicting an urbanized area of medium density may also include other semantic objects like vehicles, roads, and vegetation. The attention mechanism learns which features are relevant for each class and ranks them accordingly. The output of the attention mechanism is the weighted sum of the convolutional features extracted by the network. The weights of this function are learned in a separate step after the features have been fine-tuned. Noh~\emph{et al.}~\cite{noh2017large} formulate the attention training procedure as follows. \textcolor{black}{A scoring function $\alpha(\mathbf{f}_n;\mathbf{\Theta})$ should be learned per feature vector $ \mathbf{f}_n$, with $\mathbf{\Theta}$ representing the parameters of the function $\alpha$, $n = 1, ..., N$ the $n$-th feature vector}, and $\mathbf{f_n} \in \mathcal{R}^d$, with $d$ denoting the size of the feature vectors. Using cross-entropy as the loss, the output of the network is given by
\begin{equation}
    \mathbf{y} = \mathbf{W}(\sum_n\alpha(\mathbf{f}_n;\mathbf{\Theta})\cdot \mathbf{f}_n),
    \label{eq:multiplicative_att}
\end{equation}

with $\mathbf{W} \in \mathcal{R}^{M\times d}$ being the weights of the final prediction layer for $M$ classes. The authors use a simple CNN as the attention mechanism. This smaller network deploys the softplus~\cite{dugas2001incorporating} activation function to avoid learning negative weighting of the features. We denote this attention mechanism as multiplicative attention, since the final output is obtained by multiplication of the image features and their attention scores. The result of Equation~(\ref{eq:multiplicative_att}) completely suppresses many of the features in the image if the network does not consider them sufficiently relevant. As an alternative, we also study a different attention mechanism denoted as additive attention. In the case of additive attention, the output of the network is formulated as
\begin{equation}
    \mathbf{y} = \mathbf{W}(\sum_n (1 + \alpha(\mathbf{f}_n;\mathbf{\Theta}))\cdot \mathbf{f}_n).
    \label{eq:additive_att}
\end{equation}
While removal of irrelevant objects is desired, it is possible that for RS imagery, the multiplicative attention mechanism also removes some useful information about the context of different semantic objects. An attention mechanism that enhances the relevant features, while preserving information about their surroundings, may prove useful in the RSIR task. We propose to use additive attention, which strengthens the salient features without neutralizing all other image features, unlike multiplicative attention does. 

\begin{figure}[H]
\centering
\includegraphics[width=\textwidth]{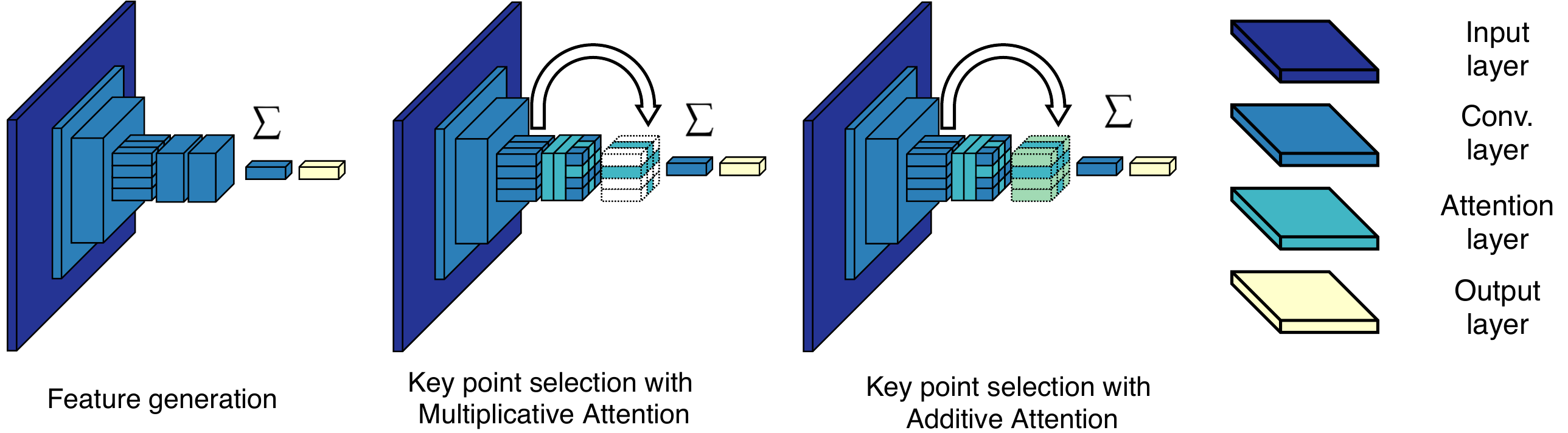}
\caption{Architectures of the feature extraction and the attention modules. Adapted from~\cite{noh2017large}.}
\label{fig:att_diag}
\end{figure}

The network architectures for feature extraction, multiplicative attention, and additive attention modules are depicted in Figure~\ref{fig:att_diag}.
For training, we follow a similar procedure as in~\cite{noh2017large}. The feature extraction network is trained as a classifier using cross-entropy as the loss. The images are resized to $224 \times 224$ pixels and used for training. We do not take random crops at this stage as the objective is to learn better features for RS images instead of feature localization and saliency. The attention modules are trained in a semi-supervised manner. In this case, the input images are randomly cropped and the crops are resized to $224 \times 224$ pixels. Cross-entropy remains as the loss to train the attention. In this way, the network should learn to identify which regions and features are more relevant for specific classes. This procedure requires no additional annotations, since only the original image label is used for training. Another advantage of this architecture is the possibility of extracting multi-scale features in a single pass. By resizing the input images, it is possible to generate features at different scales. Furthermore, manual image decomposition into patches is not required. The authors of~\cite{noh2017large} observe better retrieval performance, when the feature extraction network and the attention module are trained separately. The feature extraction network is trained first as in~\cite{noh2017large}. Then, the weights of that network are reused to train the attention network. 

After training, the features are generated for each image by removing the last two convolutional layers. Multi-scale features are extracted by resizing the input image and passing it through the network. As in the original paper, we use 7~different scales starting at 0.25 and ending at 2.0, with a separation factor of~$\sqrt{2}$. The features produced by the network are vectors of size 1024. In addition to the convolutional outcome, each DELF descriptor includes its location in the image, the attention score and the scale at which the feature was found.  For descriptor aggregation, we utilize the most attentive features only. We store 300~features per image, ranked by their attention score and across all 7~scales. \textcolor{black}{This corresponds to roughly 21\% of all the extracted descriptors with non-zero attention score. Note that features are not equally distributed across scales. For example, on the UCMerced dataset, 53\% of the features belong to the third and fourth scales (1.0 and 1.4142).} The remaining features are not exploited. 

\subsection{Descriptor aggregation}
DELF features are local, meaning that direct feature matching is computationally expensive. Each query image may contain a large number of local features that should be matched against an equally large number of features in each database image. A common solution to this problem in image retrieval literature is descriptor aggregation. This produces a compact, global representation of the image content. The most common aggregation method in RSIR literature is BoW. However, modern image retrieval systems apply other aggregation methods such as FK~\cite{perronnin2007fisher}, VLAD~\cite{jegou2010aggregating} or Aggregated Selective Match  Kernels~\cite{tolias2013aggregate}. In this work, we use VLAD to produce a single vector that encodes the visual information of the images because it outperforms both BoW and FK~\cite{jegou2010aggregating}. 

\begin{figure}[H]
\centering
\includegraphics[width=\textwidth]{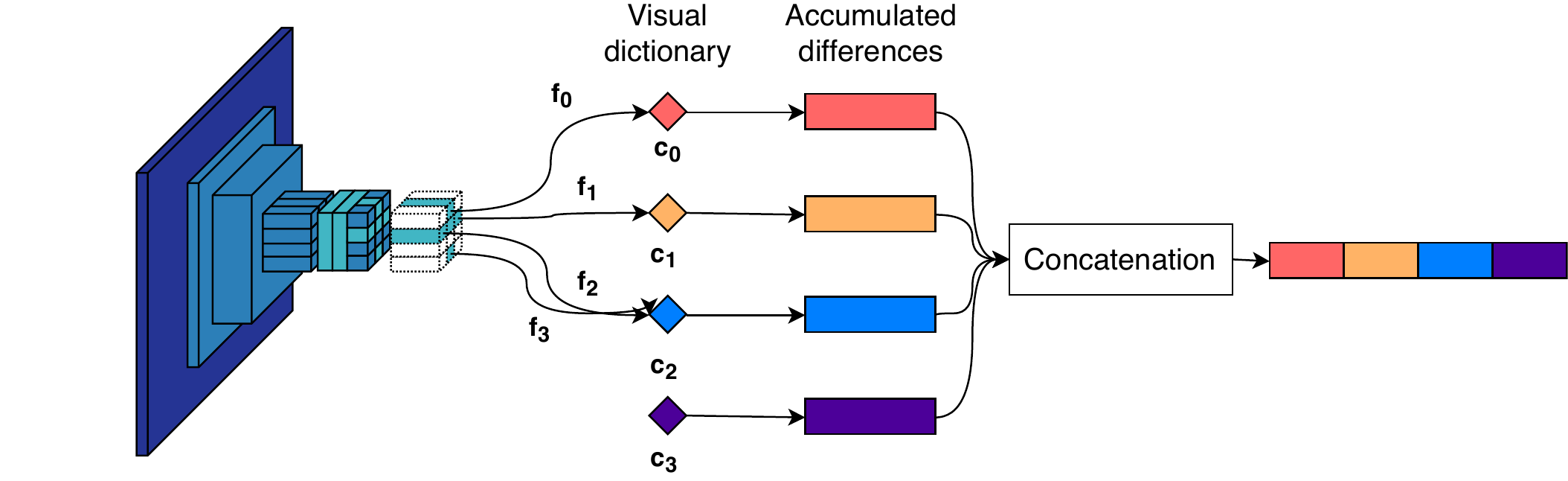}
\caption{VLAD descriptor aggregation example for $k=4$. The difference between the visual words $\mathbf{c}_i$ and the closest image features $\mathbf{f}_j$ are accumulated into $k$ vectors. Each vector is concatenated and produces a global descriptor.}
\label{fig:vlad}
\end{figure}

Similar to BoW, a visual dictionary or codebook with $k$ visual words is constructed using the k-means algorithm. The codebook $\mathbf{C}_k = \{\mathbf{c}_1, \mathbf{c}_2,...,\mathbf{c}_k\}$ is used to partition the feature space and accumulate the statistics of the local features related to the cluster center. Consider the set of $d$ dimensional local features $ F=\{\mathbf{f}_1,\mathbf{f}_2,...,\mathbf{f}_n\}$ belonging to an image. Per visual word $\mathbf{c}_i$ the difference between the features associated with $\mathbf{c}_i$ and the visual word will be computed and accumulated. The feature space is partitioned into $k$ different Voronoi cells, producing subsets of features $ F_i=\{\mathbf{f}_j; \forall j \in NN(\mathbf{c}_i) \}$, where $NN(\mathbf{c}_i)$ denotes the features closest to the visual word $\mathbf{c}_i$. The VLAD descriptor $\mathbf{v}$ is constructed as the concatenation of $k$ components, each computed by
\begin{equation}
    \mathbf{v}_i = \sum_j \mathbf{f}_j - \mathbf{c}_i,
    \label{eq:vlad_vec}
\end{equation}
with $\mathbf{v}_i$ being the vector corresponding to the $i$-th visual word resulting in a $\mathbf{d} = k \times d$ dimensional vector. The final descriptor $\mathbf{v}$ is $L_2$-normalized. The VLAD aggregation process is visualized in Figure~\ref{fig:vlad}.

An important difference between the BoW and VLAD representations is that the latter deploys visual dictionaries with relatively few visual words. In~\cite{jegou2010aggregating}, the visual dictionaries for FK and VLAD descriptors contain either 16 or 64 visual words. Meanwhile, BoW-based systems require significantly larger values of $k$. In~\cite{bai2014bag} and~\cite{tang2018unsupervised}, 50,000 and 1,500 visual words are utilized for the aggregated descriptor, respectively. The final VLAD descriptor size is dependent on both the number of visual words $k$ and the original descriptor dimension $d$. In the case of our system, the extracted local descriptors are of size 1,024. To avoid extremely large descriptors, we limit the value of $k$ to 16~visual words, resulting in VLAD vectors of size 16K. 

Regarding the training of our codebook, we use only a subset of the extracted DELF features. As mentioned in the previous section, only 300~features are preserved per image. However, training a codebook with a large number of high-dimensional descriptors is computationally expensive, even for low values of $k$. To solve this problem, we select a fraction of all available descriptors. The $N$ most attentive DELF descriptors of every image are selected for training of the visual dictionary. This results in roughly 200K to 300K descriptors per training dataset. We provide specific values of the $N$ most attentive descriptors in the next section. \textcolor{black}{Although it is possible to train the feature extraction module on other datasets, both the codebook and feature extraction module are trained individually per dataset. Thus, we obtain the most descriptive representations possible.} Once the codebook training is complete, a VLAD descriptor is generated per image. These descriptors are used in the matching module to identify visually similar images. The feature extraction and descriptor aggregation processes are performed offline for all images. 

\subsection{Image matching and query expansion}

Once each database image has a VLAD representation, it is possible to perform CBIR. Given a query image, we employ a similarity metric between its descriptor and the descriptor of every database image. The database images are ranked based on the metric and the most similar to the query image are returned to the user. The similarity is determined by the Euclidean distance of the query and database vectors.  

As mentioned previously, conventional image retrieval uses specialized techniques to improve the quality of a candidate set, with two popular techniques being geometric verification and query expansion. Geometric verification is commonly performed when positions of local features are available, and consists of matching the features found in the query image against those of the retrieved set. However, geometric verification is difficult in RS imagery, due to the significant appearance variations present in Land Use/Land Cover datasets. Alternatively, geometric verification is possible in traditional image retrieval, because transformations can be found between the local features of the query and those of the candidate images. For example, consider images of the same object or scene from different views or with partial occlusions. In these cases, a number of local descriptors will be successfully matched between the query and retrieved images. In the datasets used in this work, images belonging to the same semantic class were acquired at different geographical locations. Thereby, geometric verification becomes impossible, since the features do not belong to the same objects. Furthermore, geometric verification is computationally expensive and only a few candidates can be evaluated in this manner. The other technique, query expansion, enhances the features of the query with the features of correct candidates. In image retrieval, this is commonly done by expanding the visual words of the query with visual words of correct matches and then repeat the query. However, good matches are generally identified using geometric verification. As mentioned earlier, geometric verification is difficult to perform in RS imagery. Nevertheless, in our system, we propose to exploit a query expansion method that also has a low complexity.  

After retrieval, we select the top-three candidate images and combine their descriptors to improve the image matching process. We assume that the top-three images will consistently provide the visually closest matches and the most relevant features for query expansion. This procedure would normally require access to the local descriptors of the images to be aggregated. However, this means additional disc access operations are necessary, since local features are not loaded into memory during retrieval. To avoid this, we exploit the properties of VLAD descriptors and use Memory Vector (MV) construction to generate the expanded query descriptor. MV is a global descriptor aggregation method, which produces a group descriptor. Unlike visual words, these representations have no semantic meaning. A difference between VLAD and MV is that the latter does not require learning a visual dictionary. 

Let us consider now to combine various VLAD global descriptors into the MV representation. The purpose is to illustrate how the MV representation can prevent accessing the local descriptors of the images that expand the query.  Iscen~\emph{et al.}~\cite{iscen2017memory} have proposed two different methods for creation of MV representations. The first is denoted by \textit{p-sum} and is defined as
\begin{equation}
    \mathbf{m}(\mathbf{G}) = \mathbf{G}\mathbf{1_n},
\end{equation}
with $\mathbf{G}$ being a $m \times n$ matrix of global representations to be aggregated, where $m$ is the number of global representations and $n$ is their dimensionality. Although naive, this construction is helpful when used with VLAD descriptors. It should be born in mind that each VLAD descriptor $\mathbf{v}$ is constructed as the concatenation of $k$ vectors generated using Equation~(\ref{eq:vlad_vec}). Assume that $\mathbf{p}$ and $\mathbf{q}$ are the VLAD representations of two different images, and the row elements of the descriptor matrix $\mathbf{G}$. The \textit{p-sum} vector is given by
\begin{equation}
    \mathbf{m}(\mathbf{G}) = \mathbf{p} + \mathbf{q} = \sum_k \mathbf{p}_k + \mathbf{q}_k,
    \label{eq:psum_vlad}
\end{equation}
with $\mathbf{p}_k$, $\mathbf{q}_k$ the individual VLAD vector components of $\mathbf{p}$ and $\mathbf{q}$, respectively. From Equation~(\ref{eq:psum_vlad}) we observe that the \textit{p-sum} representation is an approximation of the VLAD vector of the features contained in both images. Instead of generating a VLAD vector from the union of the features belonging to the query and its top-three matches, we can produce an approximated representation by exploiting MV construction. The results of direct VLAD aggregation and \textit{p-sum} aggregation are not identical, because the values used in the latter have been $L_2$-normalized. However, this approximation allows us to use only the global representations of images. The second construction method, denoted \textit{p-inv}, is given by
\begin{equation}
    \mathbf{m}^+(\mathbf{G}) = (\mathbf{G^+})^T\mathbf{1_n},
\end{equation}
where $(\mathbf{G^+})$ is the Moore-Penrose pseudo-inverse~\cite{rao1972generalized} of the descriptor matrix $\mathbf{G}$. This construction down-weights the contributions of repetitive components in the vectors. In this work, we perform query expansion with both MV construction methods and compare their performance. 

\section{Experiments and discussion}
\label{s4}

This section details the applied datasets, the performed experiments and utilized hardware. First, the datasets are presented with training parameters and assessment criteria. Second, the performance of our proposed system is evaluated, using features learned for landmark recognition. The pre-trained DELF network is used for feature extraction. Third, retrieval experiments are repeated after training the feature extraction network on RS imagery. These experiments also include the comparison of two different attention mechanisms. For the previous experiments, we provide retrieval results with and without query expansion. Finally, the impact of descriptor dimensionality on the retrieval performance and image matching time is studied. The experiments are performed on a computer equipped with a GTX Titan~X GPU, a Xeon E5-2650 CPU and 32~GB of RAM.

\subsection{Experiment Parameters}
\subsubsection{Datasets and training}
Four different datasets are applied to evaluate the retrieval performance of the proposed system under different conditions. The datasets are (1) UCMerced Land Use~\cite{yang2010bag}, (2) Satellite Remote Sensing Image Database~\cite{tang2017two}, (3) Google Image Dataset of SIRI-WHU~\cite{zhao2016dirichlet,zhao2016fisher,zhu2016bag} and (4) NWPU-RESISC45~\cite{cheng2017remote}. The specifics of each dataset (image size, number of images, etc.) are found in Table~\ref{tab:data_details}. To avoid overfitting of the feature extraction network, the datasets are split. Only 80\% of the dataset images are used for training. Images belonging to the SIRI dataset are resized to $256 \times 256$ pixels to produce activation maps of the same size (height and width) as the other datasets. Regarding the  generation of the visual dictionary, only a fraction of the total features are used. In the case of UCM, SATREM, and SIRI, the first 100~features are used per image. This produces a sufficiently large feature set for codebook generation, while limiting the computational cost. The NWPU visual dictionary is trained with significantly fewer features. Here, we use only the top-10 features per image, given the much larger quantity of images contained in this dataset. As mentioned in the previous section, we select the most attentive features for each image. Both the feature extraction and visual dictionary are re-trained for each dataset. 

\begin{table}[H]
\caption{Details of the datasets used for RSIR.}
\label{tab:data_details}
\centering
\begin{tabular}{lcrccc}
\toprule
\textbf{Dataset} & \textbf{Abbreviation} & \textbf{\# of Images} & \textbf{\# of Classes} & \textbf{Image resolution} & \textbf{Spatial resolution} \\
\midrule
UCMerced \cite{yang2010bag} & UCM          & 2,100         & 21            & $256\times256$ pix.         & 0.3m                    \\
Satellite Remote \cite{tang2017two} & SATREM       & 3,000         & 20            & $256\times256$ pix.          & 0.5m                    \\
SIRI-WHU \cite{zhao2016dirichlet, zhao2016fisher, zhu2016bag}         & SIRI         & 2,400         & 12            & $200\times200$ pix.   & 2.0m                      \\
NWPU-RESISC45 \cite{cheng2017remote}    & NWPU         & 31,500        & 45            & $256\times256$ pix.          & 30 - 0.2m     \\    
\bottomrule
\end{tabular}
\end{table}

\subsubsection{Assessment criteria}

We report on the retrieval precision, since this is a common metric in image retrieval tasks. An image is considered to be a good match to the query image, if it belongs to the same semantic class as the query. Furthermore, we only consider the best 20 candidates per query to limit the number of outputs that a human would have to process. Retrieval precision is defined as the number of matches from the same class, divided by the total number of retrieved candidates. 

\subsubsection{Comparison to other RSIR systems}

The work of Tang~\emph{et al.}~\cite{tang2018unsupervised} provides a comprehensive comparison of various RSIR systems. The details of these systems are found in Table~\ref{tab:systems}. We use their obtained results as the experimental reference for our work.
\textcolor{black}{Additionally, we perform RSIR using features extracted from two state-of-the-art RS classification networks. These are Bag-of-Convolutional-Features (BoCF) \cite{cheng2017RemoteSI} and Discriminative CNNs (D-CNN) \cite{cheng2018deep}. For BoCF, we directly use the histogrammic image representations extracted from the pretrained VGG16 network. The codebook size for this method is of 5,000 visual words. Meanwhile, the D-CNN features are extracted from VGG16 fine-tuned on the corresponding dataset employing the same splits used for training DELF. For D-CNN, the parameters $\tau$, $\lambda_1$ and $\lambda_2$ are set to $0.44$, $ 0.05 $ and $0.0005$ respectively. The backbone networks and the parameters are chosen based on their performance.} We denote the results of our system without query expansion (QE) as V-DELF. As mentioned above, the expanded query vector can be produced using one of two different methods. We denote the results of the proposed system using QE for each method as either QE-S or QE-I, referring to the \textit{p-sum} or \textit{p-inv} construction, respectively.  

\begin{table}[H]
\caption{RSIR system details.}
\label{tab:systems}
\centering
\begin{tabular}{llrl}
\toprule
\textbf{Abbreviation} & \textbf{Features} & \textbf{Descriptor Size} & \textbf{Similarity Metric}\\
\midrule
RFM\cite{tang2017sar} & Edges and texture & - & Fuzzy similarity\\
SBOW\cite{yang2013geographic} & SIFT+BoW & 1,500 & $L_1$ norm\\
Hash\cite{demir2016hashing}& SIFT+BoW & 16 & Hamming dist.\\
DN7\cite{marmanis2016deep}& Convolutional & 4,096 & $L_2$ norm\\
DN8\cite{marmanis2016deep}& Convolutional & 4,096 & $L_2$ norm\\
DBOW\cite{tang2018unsupervised}& Convolutional+BoW & 1,500 & $L_1$ norm\\
BoCF\cite{cheng2017RemoteSI} & Convolutional+BoW & 5,000 & $L_2$ norm \\
DCNN\cite{cheng2018deep}& Convolutional & 25,088 & $L_2$ norm \\
ResNet-50\cite{He_2016_CVPR}& Convolutional+VLAD & 16,384\\
V-DELF (Ours) & Convolutional+VLAD & 16,384 & $L_2$ norm \\
\bottomrule
\end{tabular}
\end{table}

\subsection{RSIR using the pre-trained DELF network for feature extraction}

In this initial experiment, we deploy the pre-trained DELF network for feature extraction. This network was trained on landmark images, which are intrinsically different from RS imagery. This RSIR experiment using features trained on a different type of images, serves as a preliminary test of the overall system. Features learned for different tasks have been successfully used for image retrieval and classification tasks. Furthermore, the experiment provides insight into the descriptive power of local descriptors aggregated into VLAD vectors. Here, only the original multiplicative attention mechanism is evaluated and we utilize a visual dictionary of 16~words with a feature vector size of 1,024. 

The results of this experiment are given in Table~\ref{tab:res_trans}. From this table, it can be observed that the proposed method is capable of obtaining competitive retrieval results, outperforming most of the other RSIR systems with the exception of DBOW. Recall that the features learned by the network come from a different domain (street-level landmark images). The only part of the system that is tuned to the data statistics is the visual dictionary used for constructing the VLAD representation. These preliminary results already reveal the powerful representation provided by combining local DELF features with the VLAD descriptor, as it scores better than all other methods except for DBOW. Furthermore, we notice that the addition of query expansion (QE) consistently increases retrieval performance. Although of low complexity, our QE method yields an average performance increase of roughly 3\%. However, the gains provided by QE are class-dependent. For example, in the case of the UCM dataset, retrieval precision increases by a mere 0.3\% for the ``Harbor'' class, whereas the ``Tennis Court'' class sees an increase of roughly 7\%. Classes with good retrieval results obtain smaller gains in precision than difficult classes. This indicates that QE can have a noticeable impact on classes with poor retrieval performance. Furthermore, we observe no large differences between aggregation based on the \textit{p-sum} or \textit{p-inv} constructions.
\begin{table}[H]
\centering
\caption{Average precision comparison between various RSIR methods and DELF features (pre-trained on non-RS images).}
\begin{tabular}{lcccccccccc}
\toprule    
\textbf{Dataset} & \textbf{RFM} & \textbf{SBOW} & \textbf{Hash} & \textbf{DN7} & \textbf{DN8} & \textbf{BoCF} &\textbf{DBOW} &\textbf{V-DELF} &\textbf{ QE-S }& \textbf{QE-I} \\
 & \cite{tang2017sar} & \cite{yang2013geographic} & \cite{demir2016hashing} & \cite{marmanis2016deep} & \cite{marmanis2016deep} &  \cite{cheng2017RemoteSI} & \cite{tang2018unsupervised} & & & \\
\midrule
UCM & 0.391 & 0.532 & 0.536 & 0.704 & 0.705 & 0.243 &\textbf{0.830} & 0.746 & 0.780 & 0.780 \\
SATREM & 0.434 & 0.642 & 0.644 & 0.740 & 0.740 & 0.218 &\textbf{0.933} & 0.839 & 0.865 & 0.865 \\
SIRI & 0.407 & 0.533 & 0.524 & 0.700 & 0.700 & 0.200 &\textbf{0.926} & 0.826 & 0.869 & 0.869 \\
NWPU & 0.256 & 0.370 & 0.345 & 0.605 & 0.595 & 0.097 &\textbf{0.822} & 0.724 & 0.759 & 0.757 \\
\bottomrule
    \end{tabular}
    \label{tab:res_trans}
\end{table}

A large gap in retrieval precision is noticeable across systems employing handcrafted features versus those employing convolutional features. The latter are performing better in every dataset. Therefore, we will only compare the results of the following experiments against the systems based on convolutional features. In the next experiment, we re-train the DELF network on RS imagery and repeat this experiment. 

\subsection{RSIR with trained features}

For this experiment, we re-train the DELF network with the methodology described in the previous section. We train both the feature extraction and attention networks (multiplicative and additive) on RS imagery. New local descriptors are extracted per image and aggregated by VLAD, producing two separate descriptors, one per attention mechanism. Then, the retrieval experiments are repeated, and the retrieval precision is reported for each attention mechanism individually. \textcolor{black}{A comparison on retrieval performance to dense features extracted from fine-tuned ResNet50 is also performed. Each image is represented by $196 \times 1024$ descriptors. Considering that the dense features do not have an associated relevance score, we deploy a slightly different procedure for descriptor aggregation. First, for dictionary generation, 100 descriptors are randomly selected from the feature map. Second, the VLAD vectors are computed from all feature vectors of the dense activation maps. This process is repeated five times and the highest score is reported. The experiment is done both with features extracted at a single scale and at various scales. A reduction in performance of approximately 3\% to 4\% is observed when using multi-scale features. Therefore, we report on results using single-scale dense features. }

\begin{figure}[H]
\centering
\includegraphics[width=0.95\textwidth]{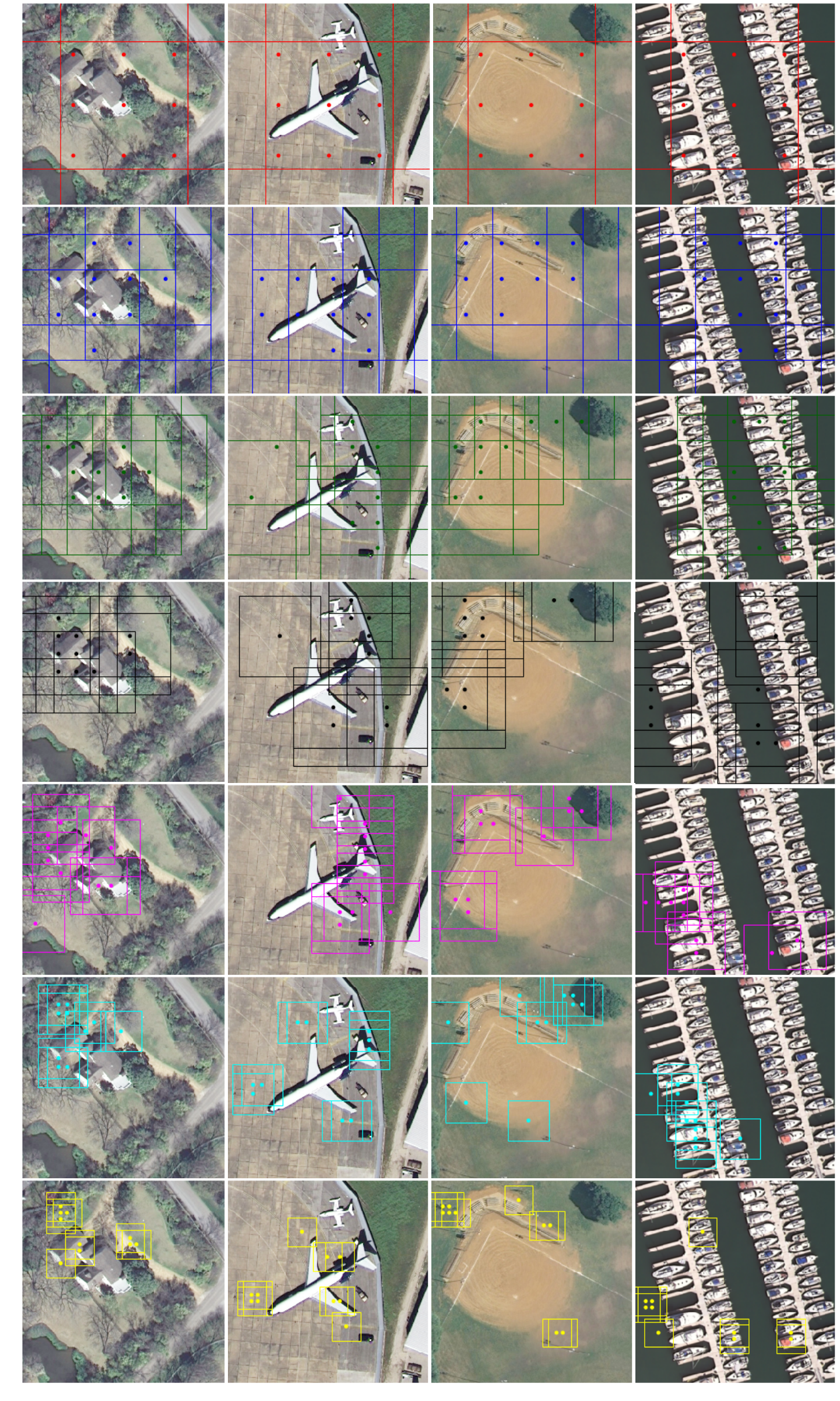}
\caption{The 10 most attentive features per scale are depicted for different semantic classes of UCM. From left to right ``Sparse residential'', ``Airplane'', ``Baseball diamond'', ``Harbor''. Each color represents a different scale. Best seen in color.}
\label{fig:receptiv}
\end{figure}

After performing several experiments on different classes, we have computed the averages of each individual experiment and captured those averages into a joint table. The results of this overall experiment are found in Table~\ref{tab:res_train}, which presents the averages of the aforementioned class experiments. The detailed precision results of each individual class are provided in Tables~\ref{tab:ir_nwpu}-\ref{tab:ir_siri}. \textcolor{black}{The attention mechanisms provide a clear gain in retrieval performance. Both attention mechanisms outperform the dense features extracted from ResNet50 with the exception of the SIRI dataset. The most likely reason for this disparity in performance across datasets is found in the training of the attention mechanism. For example, the classes with the worst performance in the SIRI dataset are ``River'' and ``Pond''. A closer inspection of the retrieval results of these classes has revealed that the most common (erroneous) retrieval results for ``River'' images are ``Pond'', ``Harbor'' and ``Water''. Meanwhile, the system confuses images in the ``Pond'' class with images from the ``River'', ``Meadows'', and ``Overpass'' classes. This indicates that the attention mechanism is not learning correctly which features are relevant for these classes, or that the features learned by the classification network are not sufficiently descriptive for some of them. Early termination of the training procedure may yield an improved performance on this disparity aspect. Figure~\ref{fig:receptiv} depicts the most attentive features per scale, and their receptive fields for different images.} We observe that for the UCM and NWPU datasets, the presented system outperforms the DBOW system by 8.59\% and 3.5\%, whereas for the SATREM and SIRI datasets the retrieval precision of the DBOW system is 3.82\% and 8.83\% higher, respectively. Summarizing, our proposal performs well on two datasets, whereas it is lower for two other datasets.  \textcolor{black}{An interesting result is that BoCF has poor overall performance for retrieval. A reasonable explanation for this behavior is the lack of training of the feature extraction network. The features extracted from the pre-trained VGG16 network are not as discriminative as those of networks trained on RS data. Meanwhile, features extracted from D-CNN provide good retrieval performance, outperforming DBOW on the UCM dataset. This performance increase is obtained through the metric loss that D-CNN employs. Metric learning produces features with low intra-class and high inter-class difference, facilitating the identification of correct candidates. Combining metric learning with deeper or attentive networks may further improve retrieval performance.}

\begin{table}[H]
\centering
\caption{Average precision comparison between various RSIR methods using convolutional features. DELF features are extracted using Multiplicative Attention (MA) and Additive Attention (AA).}
\begin{tabular}{lllll}
\toprule
 Method & UCM & SATREM & SIRI & NWPU \\
\midrule
DN7\cite{marmanis2016deep} & 0.704 & 0.740 & 0.700 & 0.605 \\
DN8\cite{marmanis2016deep} & 0.705 & 0.740 & 0.696 & 0.595 \\
DBOW\cite{tang2018unsupervised} & 0.830 & 0.933 & 0.926 & 0.821 \\
D-CNN\cite{cheng2018deep} & 0.874 & 0.852 & 0.893 & 0.736\\
ResNet50\cite{He_2016_CVPR} & 0.816 & 0.764 & 0.862 & 0.798 \\
V-DELF (MA) & 0.896 & 0.876 & 0.813 & 0.840 \\
QE-S (MA) & 0.916 & 0.895 & 0.838 & 0.857 \\
QE-I (MA) & 0.915 & 0.894 & 0.838 & 0.855 \\
V-DELF (AA) & 0.883 & 0.866 & 0.791 & 0.838 \\
QE-S (AA) & 0.854 & 0.840 & 0.818 & 0.856 \\
QE-I (AA) & 0.904 & 0.894 & 0.817 & 0.855 \\
\bottomrule
\end{tabular}
    \label{tab:res_train}
\end{table}

\begin{table}[H]
\caption{Retrieval precision for the UCM dataset after training the DELF network. Bold values indicate the best performing system. DELF descriptors are of size 16K.}
\label{tab:ir_ucm}
\centering
\begin{tabular}{lcccccccccc}
\toprule
 & DBOW & ResNet-50 & BoCF & DCNN & \multicolumn{3}{c}{Multiplicative Attention} & \multicolumn{3}{c}{Additive Attention} \\
 &\cite{tang2018unsupervised} & \cite{He_2016_CVPR} & \cite{cheng2017RemoteSI} & \cite{cheng2018deep} & V-DELF & QE-S & QE-I & V-DELF & QE-S & QE-I \\
 \midrule
Agriculture & 0.92 & 0.85 & 0.88 & \textbf{0.99} & 0.75 & 0.80 & 0.80 & 0.78 & 0.78 & 0.83 \\
Airplane & 0.95 & 0.93 & 0.11 & 0.98 & 0.98 & 0.97 & 0.97 & 0.98 & 0.94 & \textbf{0.99} \\
Baseball & \textbf{0.87} & 0.73 & 0.13 & 0.82 & 0.74 & 0.77 & 0.77 & 0.71 & 0.70 & 0.75 \\
Beach & 0.88 & \textbf{0.99} & 0.17 & \textbf{0.99} & 0.93 & 0.94 & 0.94 & 0.90 & 0.89 & 0.95 \\
Buildings & \textbf{0.93} & 0.74 & 0.10 & 0.80 & 0.83 & 0.85 & 0.85 & 0.81 & 0.80 & 0.85 \\
Chaparral & 0.94 & 0.95 & 0.93 & 1.00 & 1.00 & \textbf{1.00} & \textbf{1.00} & 0.98 & 0.94 & 0.99 \\
Dense & \textbf{0.96} & 0.62 & 0.07 & 0.65 & 0.89 & 0.90 & 0.90 & 0.87 & 0.84 & 0.89 \\
Forest & \textbf{0.99} & 0.87 & 0.78 & \textbf{0.99} & 0.97 & 0.98 & 0.98 & 0.98 & 0.94 & \textbf{0.99} \\
Freeway & 0.78 & 0.69 & 0.09 & 0.92 & 0.98 & \textbf{0.99} & \textbf{0.99} & 0.97 & 0.93 & 0.98 \\
Golf & \textbf{0.85} & 0.73 & 0.08 & 0.93 & 0.80 & 0.83 & 0.83 & 0.81 & 0.79 & \textbf{0.85} \\
Harbor & 0.95 & 0.97 & 0.20 & \textbf{1.00} & \textbf{1.00} & \textbf{1.00} & \textbf{1.00} & \textbf{1.00} & 0.95 & \textbf{1.00} \\
Intersection & 0.77 & 0.81 & 0.16 & 0.79 & 0.83 & \textbf{0.86} & 0.85 & 0.81 & 0.79 & 0.84 \\
Medium-density & 0.74 & 0.80 & 0.07 & 0.69 & 0.88 & \textbf{0.92} & 0.91 & 0.88 & 0.86 & 0.91 \\
Mobile & 0.76 & 0.74 & 0.09 & 0.89 & 0.92 & \textbf{0.94} & \textbf{0.94} & 0.86 & 0.83 & 0.89 \\
Overpass & 0.86 & 0.97 & 0.09 & 0.82 & 0.99 & 0.99 & 0.99 & \textbf{1.00} & 0.95 & \textbf{1.00} \\
Parking & 0.67 & 0.92 & 0.58 & 0.99 & 0.99 & 0.99 & 0.99 & 0.99 & 0.95 & \textbf{1.00} \\
River & 0.74 & 0.66 & 0.06 & 0.88 & 0.83 & 0.87 & 0.87 & 0.83 & 0.84 & \textbf{0.89} \\
Runway & 0.66 & 0.93 & 0.20 & 0.98 & 0.98 & \textbf{0.99} & \textbf{0.99} & 0.97 & 0.93 & 0.98 \\
Sparse & 0.79 & 0.69 & 0.11 & \textbf{0.83} & 0.76 & 0.79 & 0.78 & 0.70 & 0.65 & 0.69 \\
Storage & 0.50 & 0.86 & 0.12 & 0.60 & 0.89 & \textbf{0.93} & \textbf{0.93} & 0.78 & 0.74 & 0.80 \\
Tennis & \textbf{0.94} & 0.70 & 0.08 & 0.83 & 0.89 & \textbf{0.94} & 0.93 & 0.91 & 0.89 & \textbf{0.94} \\
\midrule
Average & 0.830 & 0.816 & 0.243 & 0.874 & 0.896 & \textbf{0.916} & 0.915 & 0.883 & 0.854 & 0.904 \\
\bottomrule
\end{tabular}
\end{table}

\begin{table}[H]
\caption{Retrieval precision for the SATREM dataset after training the DELF network. Bold values indicate the best performing system. DELF descriptors are of size 16K.}
\label{tab:ir_satrem}
\centering
\begin{tabular}{lcccccccccc}
\toprule
 & DBOW & ResNet-50 & BoCF & DCNN & \multicolumn{3}{c}{Multiplicative Attention} & \multicolumn{3}{c}{Additive Attention} \\
 &\cite{tang2018unsupervised} & \cite{He_2016_CVPR} & \cite{cheng2017RemoteSI} & \cite{cheng2018deep} & V-DELF & QE-S & QE-I & V-DELF & QE-S & QE-I \\
\midrule
Agricultural & \textbf{0.97} & 0.86 & 0.33 & 0.96 & 0.87 & 0.90 & 0.90 & 0.84 & 0.81 & 0.86 \\
Airplane & \textbf{0.96} & 0.86 & 0.12 & 0.64 & 0.84 & 0.88 & 0.88 & 0.89 & 0.85 & 0.89 \\
Artificial & \textbf{0.97} & 0.93 & 0.11 & 0.92 & 0.82 & 0.81 & 0.81 & 0.90 & 0.86 & 0.91 \\
Beach & \textbf{0.95} & 0.86 & 0.07 & 0.85 & 0.84 & 0.87 & 0.87 & 0.81 & 0.78 & 0.83 \\
Building & \textbf{0.97} & 0.92 & 0.14 & 0.83 & 0.92 & 0.94 & 0.94 & 0.82 & 0.79 & 0.83 \\
Chaparral & \textbf{0.96} & 0.79 & 0.13 & 0.88 & 0.86 & 0.89 & 0.90 & 0.93 & 0.90 & 0.95 \\
Cloud & \textbf{0.99} & 0.97 & 0.80 & 1.00 & 0.96 & 0.97 & 0.97 & 0.88 & 0.85 & 0.90 \\
Container & 0.96 & 0.97 & 0.08 & 0.88 & 0.99 & \textbf{1.00} & \textbf{1.00} & 0.84 & 0.81 & 0.86 \\
Dense & \textbf{1.00} & 0.89 & 0.17 & 0.90 & 0.94 & 0.94 & 0.94 & 0.95 & 0.93 & 0.98 \\
Factory & 0.91 & 0.69 & 0.09 & 0.76 & 0.72 & 0.74 & 0.74 & 0.98 & 0.94 & \textbf{0.99} \\
Forest & 0.96 & 0.89 & 0.70 & 0.96 & 0.94 & 0.95 & 0.95 & 0.98 & 0.95 & \textbf{1.00} \\
Harbor & \textbf{0.98} & 0.80 & 0.11 & 0.77 & 0.95 & 0.96 & 0.96 & 0.91 & 0.87 & 0.92 \\
Medium-Density & \textbf{1.00} & 0.67 & 0.08 & 0.63 & 0.67 & 0.67 & 0.67 & 0.83 & 0.80 & 0.85 \\
Ocean & 0.92 & 0.91 & 0.85 & \textbf{0.98} & 0.91 & 0.92 & 0.92 & 0.76 & 0.74 & 0.79 \\
Parking & 0.95 & 0.87 & 0.10 & 0.87 & 0.94 & \textbf{0.96} & \textbf{0.96} & 0.88 & 0.85 & 0.90 \\
River & 0.71 & 0.83 & 0.07 & 0.80 & 0.70 & 0.74 & 0.74 & 0.99 & 0.95 & \textbf{1.00} \\
Road & 0.82 & 0.85 & 0.10 & 0.89 & 0.90 & 0.93 & 0.92 & 0.96 & 0.95 & \textbf{0.99} \\
Runway & 0.86 & 0.96 & 0.09 & 0.96 & 0.95 & \textbf{0.97} & \textbf{0.97} & 0.81 & 0.79 & 0.84 \\
Sparse & \textbf{0.92} & 0.75 & 0.09 & 0.67 & 0.84 & 0.85 & 0.84 & 0.61 & 0.58 & 0.63 \\
Storage & 0.91 & 0.98 & 0.12 & 0.89 & 0.99 & \textbf{1.00} & \textbf{1.00} & 0.95 & 0.91 & 0.96 \\
\midrule
Average & \textbf{0.933} & 0.862 & 0.218 & 0.852 & 0.876 & 0.895 & 0.894 & 0.866 & 0.840 & 0.894 \\
\bottomrule
\end{tabular}
\end{table}

\begin{table}[H]
\caption{Retrieval precision for the SIRI dataset after training the DELF network. Bold values indicate the best performing system. DELF descriptors are of size 16K.}
\label{tab:ir_siri}
\centering
\begin{tabular}{lcccccccccc}
\toprule
 & DBOW & ResNet-50 & BoCF & DCNN & \multicolumn{3}{c}{Multiplicative Attention} & \multicolumn{3}{c}{Additive Attention} \\
 & \cite{tang2018unsupervised} & \cite{He_2016_CVPR} & \cite{cheng2017RemoteSI} & \cite{cheng2018deep} & V-DELF & QE-S & QE-I & V-DELF & QE-S & QE-I \\
 \midrule
Agriculture & 0.99 & 0.95 & 0.17 & \textbf{1.00} & 0.92 & 0.94 & 0.94 & 0.90 & 0.93 & 0.93 \\
Commercial & \textbf{0.99} & 0.90 & 0.17 & \textbf{0.99} & 0.95 & 0.97 & 0.97 & 0.95 & 0.97 & 0.96 \\
Harbor & \textbf{0.89} & 0.63 & 0.12 & 0.79 & 0.71 & 0.74 & 0.74 & 0.75 & 0.78 & 0.78 \\
Idle & \textbf{0.97} & 0.63 & 0.11 & 0.89 & 0.77 & 0.80 & 0.80 & 0.79 & 0.83 & 0.84 \\
Industrial & 0.90 & 0.88 & 0.14 & \textbf{0.96} & 0.92 & \textbf{0.96} & \textbf{0.96} & 0.93 & \textbf{0.96} & \textbf{0.96} \\
Meadow & \textbf{0.93} & 0.77 & 0.29 & 0.86 & 0.78 & 0.82 & 0.82 & 0.79 & 0.81 & 0.81 \\
Overpass & 0.89 & 0.80 & 0.21 & \textbf{0.95} & 0.92 & 0.94 & 0.94 & 0.92 & \textbf{0.95} & \textbf{0.95} \\
Park & 0.87 & 0.82 & 0.11 & 0.91 & 0.85 & 0.90 & 0.90 & 0.88 & \textbf{0.92} & \textbf{0.92} \\
Pond & \textbf{0.97} & 0.57 & 0.12 & 0.81 & 0.71 & 0.74 & 0.74 & 0.74 & 0.77 & 0.76 \\
Residential & \textbf{0.97} & 0.84 & 0.11 & 0.96 & 0.91 & 0.94 & 0.94 & 0.90 & 0.93 & 0.93 \\
River & \textbf{0.89} & 0.44 & 0.12 & 0.60 & 0.66 & 0.69 & 0.69 & 0.69 & 0.72 & 0.71 \\
Water & 0.86 & 0.94 & 0.72 & \textbf{1.00} & 0.98 & 0.99 & 0.99 & 0.98 & 0.99 & 0.99 \\
\midrule
Average & \textbf{0.926} & 0.764 & 0.200 & 0.893 & 0.840 & 0.869 & 0.867 & 0.851 & 0.880 & 0.879 \\
\bottomrule
\end{tabular}
\end{table}

\begin{table}[H]
\caption{Retrieval precision for the NWPU dataset after training the DELF network. Bold values indicate the best performing system. DELF descriptors are of size 16K.}
\label{tab:ir_nwpu}
\centering
\begin{tabular}{lcccccccccc}
\toprule
 & DBOW & ResNet-50 &  BoCF & D-CNN & \multicolumn{3}{c}{Multiplicative Attention} & \multicolumn{3}{c}{Additive Attention} \\
 & \cite{tang2018unsupervised} & \cite{He_2016_CVPR} & \cite{cheng2017RemoteSI} & \cite{cheng2018deep} & V-DELF & QE-S & QE-I & V-DELF & QE-S & QE-I \\
 \midrule
Airplane & \textbf{0.98} & 0.88 & 0.04 & 0.82 & 0.92 & 0.93 & 0.93 & 0.95 & 0.96 & 0.96 \\
Airport & \textbf{0.95} & 0.72 & 0.03 & 0.64 & 0.79 & 0.81 & 0.81 & 0.80 & 0.82 & 0.82 \\
Baseball Diamond & \textbf{0.86} & 0.69 & 0.04 & 0.61 & 0.65 & 0.64 & 0.64 & 0.64 & 0.61 & 0.61 \\
Basketball Court & \textbf{0.83} & 0.61 & 0.03 & 0.59 & 0.70 & 0.71 & 0.71 & 0.72 & 0.73 & 0.73 \\
Beach & 0.85 & 0.77 & 0.03 & 0.78 & 0.81 & 0.83 & 0.83 & 0.84 & \textbf{0.86} & \textbf{0.86} \\
Bridge & \textbf{0.95} & 0.73 & 0.04 & 0.79 & 0.77 & 0.81 & 0.81 & 0.78 & 0.82 & 0.82 \\
Chaparral & 0.96 & 0.98 & 0.62 & \textbf{0.99} & 0.99 & \textbf{0.99} & \textbf{0.99} & 0.98 & \textbf{0.99} & \textbf{0.99} \\
Church & \textbf{0.80} & 0.56 & 0.05 & 0.39 & 0.64 & 0.64 & 0.64 & 0.57 & 0.56 & 0.56 \\
Circular Farmland & 0.94 & 0.97 & 0.03 & 0.88 & 0.98 & \textbf{0.99} & \textbf{0.99} & 0.97 & 0.98 & 0.98 \\
Cloud & \textbf{0.98} & 0.92 & 0.06 & 0.93 & 0.97 & \textbf{0.98} & \textbf{0.98} & 0.96 & 0.97 & 0.97 \\
Commercial Area & 0.79 & 0.82 & 0.04 & 0.59 & 0.84 & \textbf{0.88} & \textbf{0.88} & 0.84 & \textbf{0.88} & \textbf{0.88} \\
Dense Residential & 0.90 & 0.89 & 0.06 & 0.76 & 0.91 & \textbf{0.95} & 0.94 & 0.87 & 0.90 & 0.90 \\
Desert & \textbf{0.97} & 0.87 & 0.30 & 0.89 & 0.91 & 0.92 & 0.92 & 0.89 & 0.91 & 0.91 \\
Forest & 0.95 & 0.95 & 0.60 & 0.94 & 0.96 & \textbf{0.97} & \textbf{0.97} & 0.96 & \textbf{0.97} & \textbf{0.97} \\
Freeway & 0.64 & 0.65 & 0.04 & 0.64 & 0.81 & \textbf{0.86} & \textbf{0.86} & 0.82 & 0.85 & 0.85 \\
Golf Course & 0.82 & 0.96 & 0.03 & 0.86 & 0.96 & \textbf{0.97} & \textbf{0.97} & 0.96 & \textbf{0.97} & \textbf{0.97} \\
Ground Track Field & \textbf{0.80} & 0.63 & 0.05 & 0.62 & 0.76 & 0.77 & 0.77 & 0.74 & 0.76 & 0.75 \\
Harbor & 0.88 & 0.93 & 0.06 & 0.90 & 0.96 & 0.97 & 0.97 & 0.97 & \textbf{0.98} & \textbf{0.98} \\
Industrial Area & 0.85 & 0.75 & 0.03 & 0.65 & 0.85 & 0.88 & 0.88 & 0.86 & \textbf{0.89} & \textbf{0.89} \\
Intersection & 0.80 & 0.64 & 0.06 & 0.58 & 0.73 & 0.72 & 0.72 & 0.83 & \textbf{0.86} & \textbf{0.86} \\
Island & 0.88 & 0.88 & 0.17 & 0.90 & 0.93 & 0.94 & \textbf{0.95} & 0.92 & 0.94 & 0.94 \\
Lake & \textbf{0.85} & 0.80 & 0.03 & 0.75 & 0.83 & 0.85 & \textbf{0.85} & 0.78 & 0.79 & 0.79 \\
Meadow & 0.90 & 0.84 & 0.63 & 0.88 & 0.91 & \textbf{0.93} & \textbf{0.93} & 0.90 & 0.92 & 0.92 \\
Medium Residential & \textbf{0.94} & 0.78 & 0.03 & 0.64 & 0.77 & 0.77 & 0.77 & 0.80 & 0.81 & 0.81 \\
Mobile Home Park & 0.83 & 0.93 & 0.04 & 0.85 & 0.96 & \textbf{0.97} & \textbf{0.97} & 0.93 & 0.95 & 0.95 \\
Mountain & 0.95 & 0.88 & 0.07 & 0.85 & 0.95 & \textbf{0.96} & \textbf{0.96} & 0.92 & 0.94 & 0.94 \\
Overpass & 0.74 & 0.87 & 0.04 & 0.67 & 0.88 & 0.90 & 0.90 & 0.88 & \textbf{0.91} & \textbf{0.91} \\
Palace & \textbf{0.80} & 0.41 & 0.04 & 0.30 & 0.53 & 0.56 & 0.56 & 0.48 & 0.48 & 0.48 \\
Parking Lot & 0.70 & 0.95 & 0.09 & 0.90 & 0.95 & \textbf{0.97} & \textbf{0.97} & 0.94 & 0.96 & 0.96 \\
Railway & 0.84 & 0.88 & 0.07 & 0.81 & 0.87 & 0.89 & 0.89 & 0.87 & \textbf{0.90} & \textbf{0.90} \\
Railway Station & \textbf{0.86} & 0.62 & 0.03 & 0.55 & 0.71 & 0.73 & 0.73 & 0.72 & 0.75 & 0.75 \\
Rectangular Farmland & 0.66 & 0.82 & 0.06 & 0.79 & 0.86 & \textbf{0.88} & \textbf{0.88} & 0.87 & \textbf{0.88} & \textbf{0.88} \\
River & 0.76 & 0.70 & 0.03 & 0.59 & 0.73 & 0.75 & 0.75 & 0.75 & \textbf{0.77} & \textbf{0.77} \\
Roundabout & 0.83 & 0.72 & 0.11 & 0.75 & 0.88 & 0.90 & 0.90 & 0.89 & \textbf{0.91} & \textbf{0.91} \\
Runway & 0.78 & 0.80 & 0.04 & 0.81 & 0.87 & \textbf{0.89} & \textbf{0.89} & 0.85 & 0.87 & 0.87 \\
Sea Ice & 0.90 & 0.98 & 0.12 & 0.97 & 0.98 & \textbf{0.99} & \textbf{0.99} & \textbf{0.99} & \textbf{0.99} & \textbf{0.99} \\
Ship & 0.65 & 0.61 & 0.06 & 0.73 & 0.65 & 0.69 & 0.69 & 0.77 & \textbf{0.79} & \textbf{0.79} \\
Snowberg & 0.83 & 0.97 & 0.04 & 0.96 & 0.97 & \textbf{0.98} & \textbf{0.98} & 0.97 & \textbf{0.98} & \textbf{0.98} \\
Sparse Residential & \textbf{0.84} & 0.69 & 0.05 & 0.71 & 0.69 & 0.70 & 0.70 & 0.76 & 0.78 & 0.78 \\
Stadium & 0.57 & 0.81 & 0.08 & 0.62 & 0.85 & \textbf{0.86} & \textbf{0.86} & 0.78 & 0.80 & 0.80 \\
Storage Tank & 0.48 & 0.88 & 0.07 & 0.86 & 0.92 & \textbf{0.94} & 0.93 & 0.89 & 0.91 & 0.91 \\
Tennis Court & 0.72 & \textbf{0.80} & 0.03 & 0.38 & 0.77 & 0.78 & 0.78 & 0.70 & 0.69 & 0.68 \\
Terrace & 0.76 & 0.88 & 0.03 & 0.83 & 0.88 & 0.90 & 0.89 & 0.89 & \textbf{0.91} & \textbf{0.91} \\
Thermal Power Station & 0.72 & 0.68 & 0.04 & 0.47 & \textbf{0.78} & \textbf{0.78} & \textbf{0.78} & 0.74 & 0.76 & 0.76 \\
Wetland & 0.70 & 0.82 & 0.08 & 0.71 & 0.77 & 0.80 & 0.79 & 0.81 & \textbf{0.83} & \textbf{0.83} \\
\midrule
Average & 0.821 & 0.798 & 0.097 & 0.736 & 0.840 & \textbf{0.857} & 0.855 & 0.838 & 0.856 & 0.855 \\
\bottomrule
\end{tabular}
\end{table}

We now discuss the difference in performance of the two attention mechanisms. It can be directly noticed from the results that the multiplicative attention provides better performance than the additive attention mechanism. For all datasets, multiplicative attention provides a higher average precision. This performance difference indicates that context information has a distracting effect on many semantic classes. Thereby, it reduces the descriptiveness of the features extracted from the network which then hinders the retrieval procedure. With respect to the MV construction method, multiplicative attention presents only minor differences between both MV construction methods, whereas additive attention results in somewhat larger variations. A clear example of this aspect, is found in the results of the UCM and NWPU datasets. The average retrieval precision when using additive attention and \textit{p-inv} construction is 5\% higher than for the \textit{p-sum} construction. This can be explained by the normalization effect of the \textit{p-inv} construction and the context information included in the additive attention mechanism. Since the \textit{p-inv} construction uses the Moore-Penrose pseudo-inverse, the contributions of repetitive or bursty features are normalized. This does not happen in the \textit{p-sum} construction, which deteriorates the performance.

\begin{figure}[H]
\centering
\includegraphics[width=\textwidth]{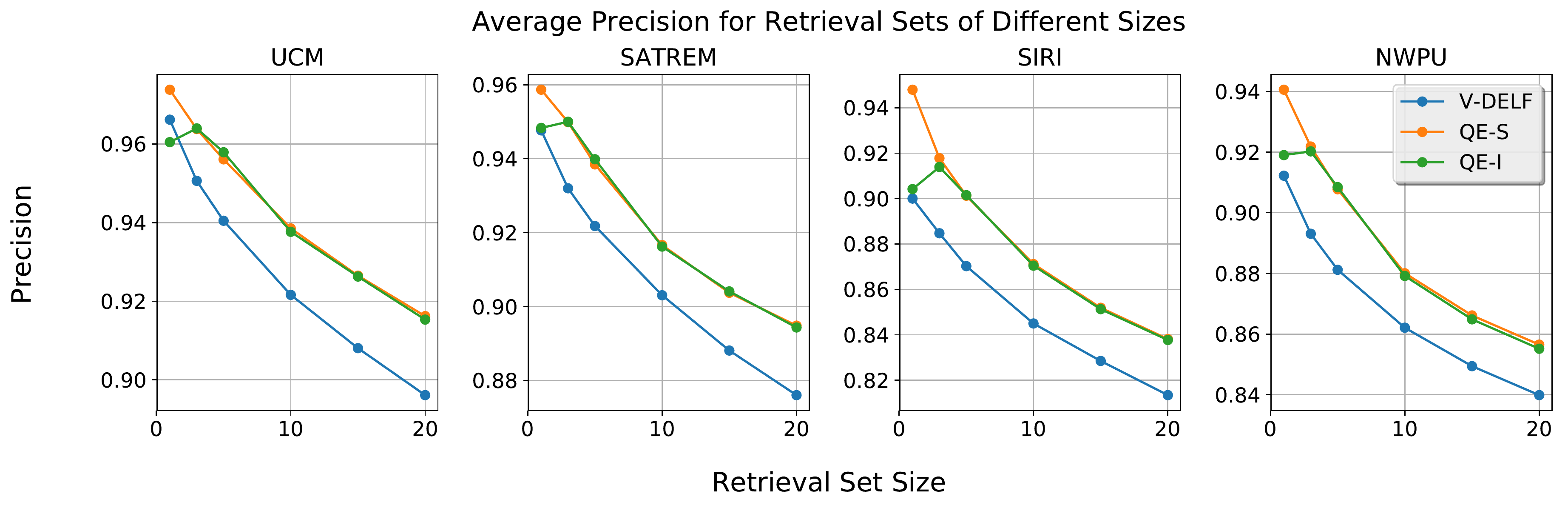}
\caption{Average precision for different sizes of the retrieval set. Average precision is reported for retrieval sets with 1, 3, 5, 10, 15, and 20 images.}
\label{fig:prec_at_N}
\end{figure}

\begin{figure}[H]
\centering
\includegraphics[width=0.75\textwidth]{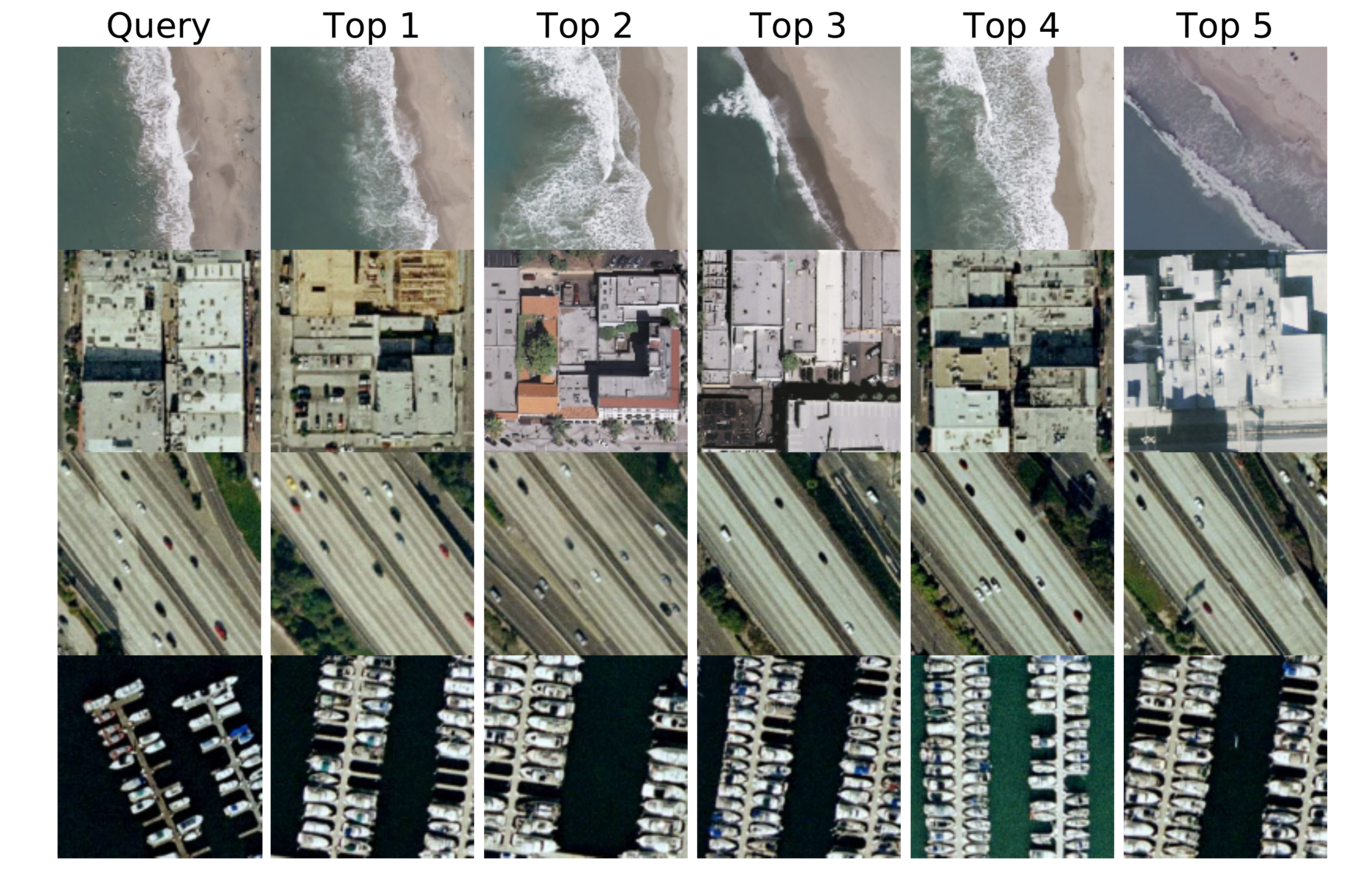}
\caption{Example queries and retrieved images for the UCM dataset. From top to bottom, the query images belong to the classes, ``Beach'', ``Buildings'', ``Freeway'' and ``Harbor''.}
\label{fig:ret_images_ucm}
\end{figure}

\begin{figure}[H]
\centering
\includegraphics[width=0.75\textwidth]{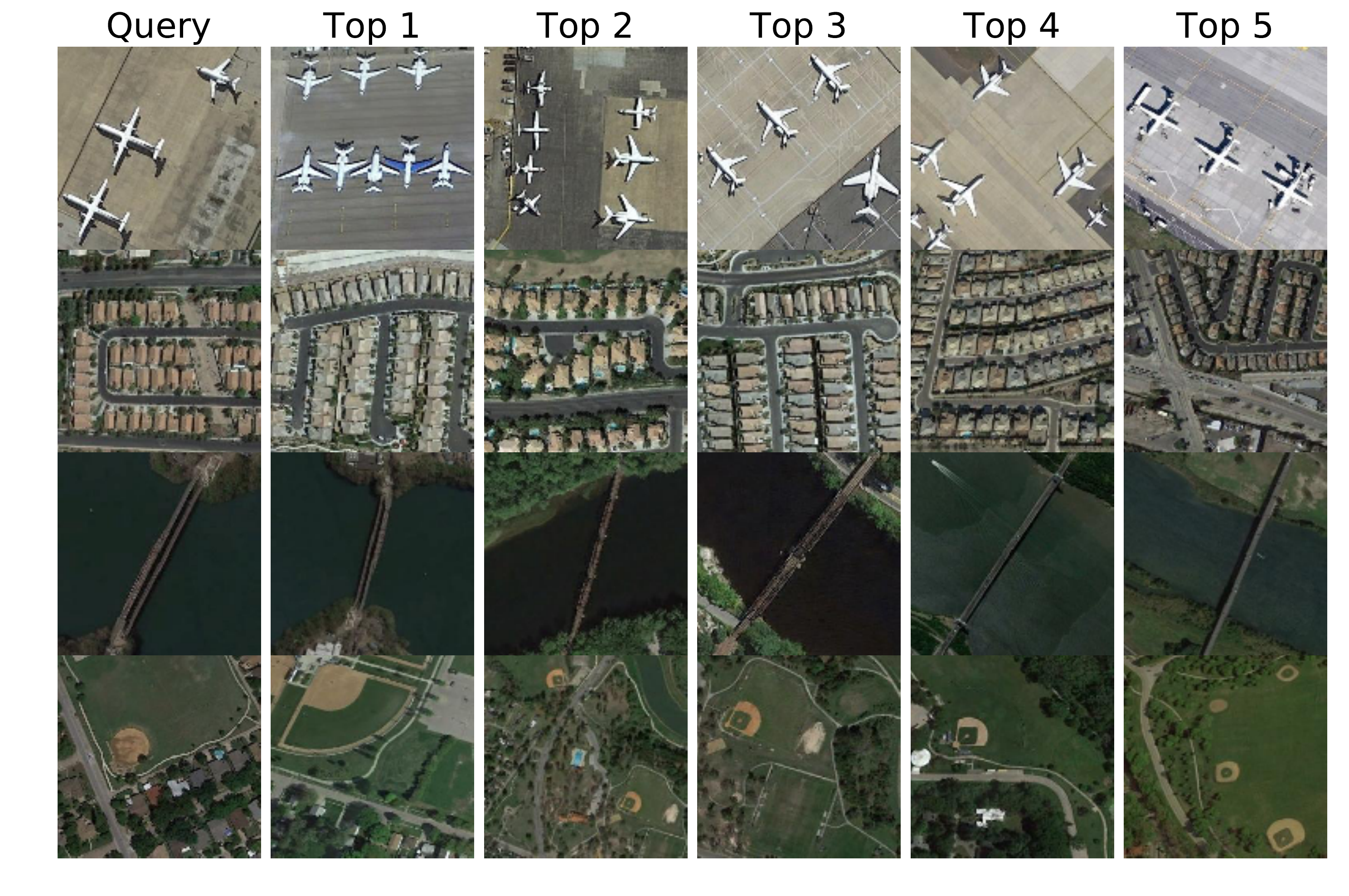}
\caption{Example queries and retrieved images for the NWPU dataset. From top to bottom, the query images belong to the classes, ``Airport'', ``Harbor'', ``Bridge'' and ``Baseball diamond''.}
\label{fig:ret_images_nwpu}
\end{figure}

A further aspect for evaluation is the query expansion stability by experimenting with different sizes of the retrieval set, using all datasets and the multiplicative attention only as it provides the best performance. In Figure~\ref{fig:prec_at_N}, the average precision for various sizes of the retrieval set is depicted. It can be observed that, on the average, retrieval performance is high when retrieving up to five images. Particularly, retrieval precision is higher than 0.88 for all datasets when considering only the top-three images. This outcome explains why our query expansion method consistently increases performance. Our assumption that good images are found in the top-three positions is indeed confirmed. In the large majority of the cases, features from those images facilitate better matching. With regards to the MV constructions for query expansion, it can be seen that query expansion always improves the results. Furthermore, their results are almost identical, with the exception of top-one precision. We also present qualitative matching results in Figures~\ref{fig:ret_images_ucm} and \ref{fig:ret_images_nwpu}, where some examples are shown of queries and retrieved images from the UCM and NWPU datasets. 

\textcolor{black}{The impact of two additional parameters on retrieval performance is evaluated. First, we study how the number of descriptors used for constructing the codebook affects the retrieval precision. For this experiment, VLAD descriptors are constructed with the 300~most attentive features and the number of features used for codebook training is changed. Second, we evaluate how system performance is affected by the number of features used when constructing the VLAD descriptors. In this case, codebooks are generated using 100~descriptors as in the previous experiments. The results of this study are found in Table \ref{tab:ablation}. The first experiment reveals that using more features for generation of the codebook is beneficial, as performance generally increases. However, when too many features are used performance may decrease, as is the case for the UCM and SIRI datsets. From the second experiment, we observe that the number of features used for VLAD generation has a much larger impact on the quality of the retrieved image sets. Using insufficient features per image leads to poor retrieval quality. This indicates that a few features (even if they are the most attentive or salient)  are not sufficient to correctly describe the image content. The retrieval precision increases when more features are used for VLAD construction. Nevertheless, performance appears to saturate around 300 descriptors.}

\begin{table}[H]
\caption{Impact of the number of descriptors used for visual dictionary and VLAD construction on retrieval performance. The number of the descriptors used for training the NWPU codebooks is one tenth of the number reported on the table.}
\label{tab:ablation}
\centering
\begin{tabular}{lcccc|ccccc}
\toprule
Dataset & \multicolumn{4}{l|}{\textbf{\# of descriptors used for codebook}} & \multicolumn{5}{l}{\textbf{\# of descriptors used for VLAD computation}} \\
 & 50  & 100 & 150 & 300 & 50 & 100 & 200 & 300 & 400 \\
\midrule
UCM & 0.893 & 0.896 & 0.897 & 0.861 & 0.588 & 0.691 & 0.826 & 0.896 & 0.906 \\
SATREM & 0.880 & 0.876 & 0.879 & 0.884 & 0.710 & 0.794 & 0.844 & 0.876 & 0.878 \\
SIRI & 0.807 & 0.813 & 0.812 & 0.792 & 0.645 & 0.714 & 0.767 & 0.813 & 0.812 \\
NWPU & 0.838 & 0.840 & 0.851 & 0.858 & 0.798 &0.832 & 0.842 & 0.840 & 0.842  \\
\bottomrule
\end{tabular}
\end{table}

While the proposed system provides state-of-the-art results in the RSIR task, the above described experiments deploy descriptors that are 4~times larger than the largest descriptor used in previous work. Since image retrieval systems should be computationally efficient, we reduce the dimensionality of our representations and study the impact on the retrieval performance.

\subsection{Dimensionality reduction}

\begin{figure}[H]
\centering
\includegraphics[width=\textwidth, height=8cm]{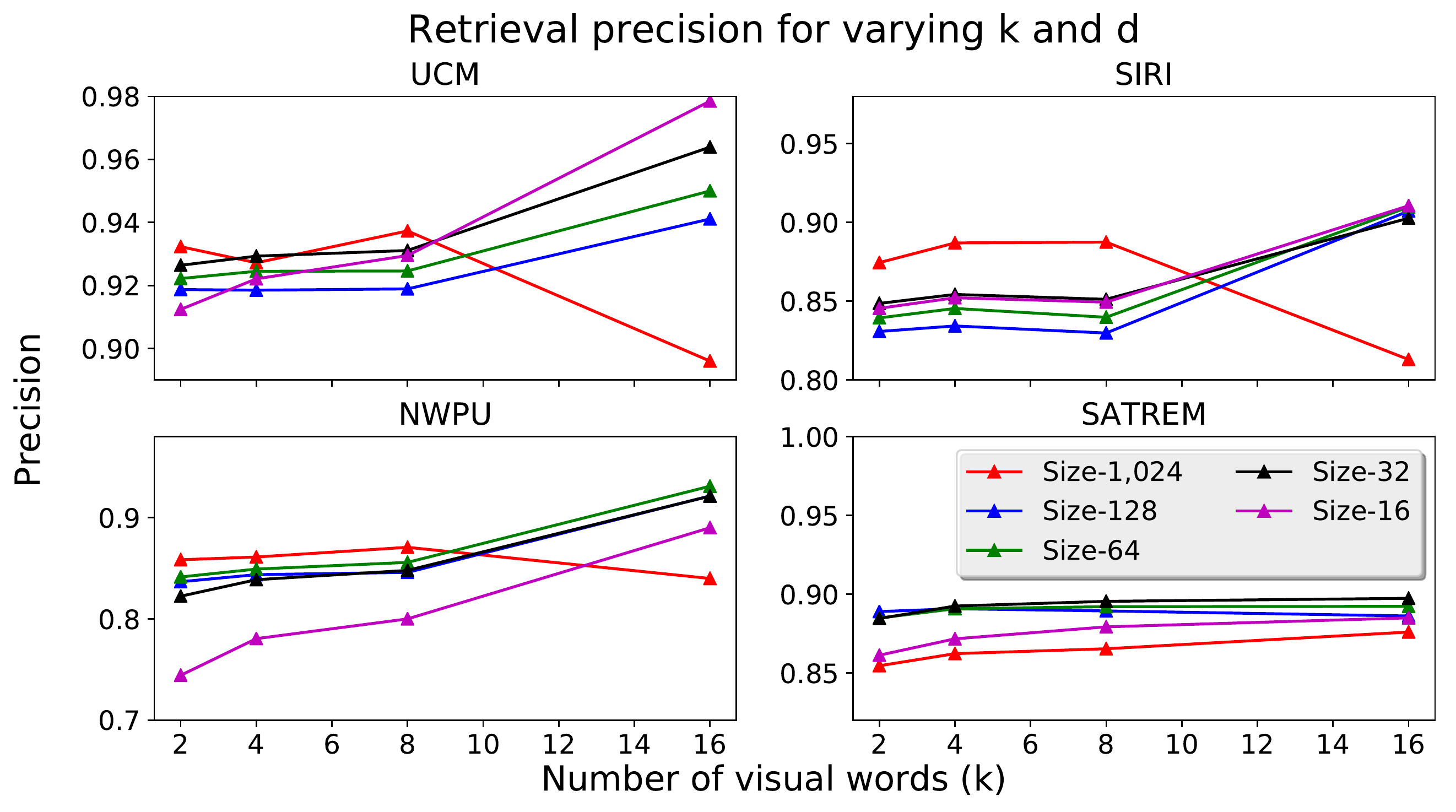}
\caption{Retrieval precision using lower dimensional descriptors for all datasets. }
\label{fig:pca}
\end{figure}

The presented system has two different parameters that affect the final descriptor dimensionality: the number of visual words in our codebook $k$ and the original feature vector size $d$. We have studied the impact of these two parameters on the retrieval precision. PCA was used to reduce the original feature vector dimensions to 1,024, 128, 64, 32 or 16 and apply 16, 8, 4 or 2 visual words. The principal components are learned using the same collection of features used for generation of the codebook. An alternative method to produce lower dimensional descriptors is to directly reduce the dimensionality of the VLAD vector. This has been demonstrated to be effective in \cite{jegou2010aggregating}, due to the sparsity and structure of the VLAD representation. However, this procedure does not reduce the computation time for codebook generation. We repeat the retrieval experiments on each dataset for each combination of descriptor size and length of the visual dictionary. The results are presented in Tables~\ref{tab:pca_avg}, \ref{tab:pca_vlad} and Figure~\ref{fig:pca}. 

We observe two interesting trends when reducing descriptor dimensionality. First, dimensionality reduction by PCA appears to be an effective way for generating concise yet highly descriptive representations. Dimensionality reduction applied to either the feature vectors or the VLAD descriptor provides performance gains. Performance is either comparable to the full-sized descriptors, or even better. This yields global representations which are smaller than those used in other work, with the exception of~\cite{demir2016hashing}. Furthermore, directly applying PCA to the feature vectors provides larger performance gains than reducing the dimensionality of the VLAD descriptor. \textcolor{black}{The likely cause for this behavior is the removal of irrelevant components present in the original feature vectors. The contributions of these components cannot be completely removed from the VLAD descriptor after aggregation.} Second, the best descriptor size $d$ varies with the datasets. Medium-sized descriptors (of size 64 or 32) with larger codebooks provide the best results and outperform the full 16K-dimensional representation. We observe that when increasing the number of visual words, also the performance increases. However, saturation occurs rapidly, so there is a limit to the performance gain introduced by larger values of $k$. The observed performance difference is due to the concentration of PCA components in each extracted feature vector. Extracting 1,024-dimensional features seems to include irrelevant data. The information contained in the feature vectors can be compressed to a certain extent, without affecting the overall descriptiveness of the vector. Since shorter descriptors provide performance gains, we compare the retrieval precision with these descriptors ($d=32$ and $d=64$ with $k=8$) against the DBOW system. The results are plotted in Table~\ref{tab:pca_avg}. From this table, it can be concluded that the gains from having adequately-sized feature vectors and visual vocabularies are significant. With smaller descriptors, we achieve even better retrieval performance, resulting in an improvement of 11.9\% and 5.7\% on the UCM and NWPU datasets w.r.t. the state-of-the-art method. Meanwhile, the earlier performance improvement in the SATREM and SIRI datasets is reduced to about half, giving 2\% and 4.6\%, respectively. 
\begin{table}[H]
\caption{Average retrieval precision using lower-dimensional descriptors. PCA is applied to DELF features.}
\label{tab:pca_avg}
\centering
\begin{tabular}{lllllllllll}
\toprule
\textbf{Dataset} &  & \multicolumn{3}{c}{$k=16$, $d=1,024$} & \multicolumn{3}{c}{$k=8$, $d=64$} & \multicolumn{3}{c}{$k=8$, $d=32$} \\
 & \textbf{DBOW}\cite{tang2018unsupervised} &\textbf{ V-DELF} & \textbf{QE-S} &\textbf{ QE-I} & \textbf{V-DELF} & \textbf{QE-S} & \textbf{QE-I} & \textbf{V-DELF} & \textbf{QE-S} & \textbf{QE-I} \\
 \midrule
UCM & 0.830 & 0.896 & 0.916 & 0.915 & 0.925 & 0.943 & 0.943 & 0.931 & \textbf{0.949} & \textbf{0.949} \\
SATREM & \textbf{0.933} & 0.876 & 0.895 & 0.894 & 0.892 & 0.910 & 0.910 & 0.897 & 0.913 & 0.913 \\
SIRI & \textbf{0.926} & 0.813 & 0.838 & 0.838 & 0.840 & 0.869 & 0.867 & 0.851 & 0.880 & 0.879 \\
NWPU & 0.821 & 0.840 & 0.857 & 0.855 & 0.856 & \textbf{0.878} & 0.876 & 0.848 & 0.873 & 0.871 \\
\bottomrule
\end{tabular}
\end{table}

\begin{table}[H]
\caption{Average retrieval precision with V-DELF using lower-dimensional descriptors. PCA is applied to reduce the size of the VLAD descriptors.}
\label{tab:pca_vlad}
\centering
\begin{tabular}{lccccccc}
\toprule
Dataset & \multicolumn{7}{c}{VLAD descriptor size} \\
 & 16,384 & 1,024 & 512 & 256 & 128 & 64 & 32 \\
 \midrule
UCM & 0.896 & 0.894 & 0.898 & 0.903 & 0.907 & \textbf{0.912} & 0.907 \\
SATREM & 0.876 & 0.876 & 0.880 & 0.884 & 0.887 & \textbf{0.889} & 0.885 \\
SIRI & 0.813 & 0.811 & 0.816 & 0.822 & 0.827 & \textbf{0.830} & 0.819 \\
NWPU & 0.840 & 0.858 & 0.862 & 0.865 & \textbf{0.866} & 0.859 & 0.818 \\
\bottomrule
\end{tabular}
\end{table}

\textcolor{black}{Finally, we  study the computational complexity of the system. First, the time required for multi-scale feature extraction is measured. On the average, obtaining the DELF descriptors from a single image requires only 20~milliseconds. A more computationally expensive operation is the construction of the codebook. K-means clustering of the selected features (with $k$=16) takes roughly 5~minutes. However, codebook generation is an off-line process and is not performed every time a query is made. If a query image is new, its VLAD representation cannot be pre-computed. As such, we have also measured the time necessary for producing VLAD descriptors, assuming that the features and the codebook are readily available. VLAD aggregation is done in approximately 10~milliseconds. This means that processing a new query has, on the average, a 30~millisecond overhead.} We also perform a comparison of the retrieval execution times. The time it takes to traverse the entire database and retrieve the images depends on two quantities. First, the size of the database, since we check against every descriptor in it. Second, the vector size, because metrics using longer vectors take more execution time. We follow the procedure of~\cite{tang2018unsupervised} and compare the results of our original and compressed descriptors for various database sizes. These results are reported in Table~\ref{tab:times}, wherein the retrieval execution times for other RSIR systems are taken from~\cite{tang2018unsupervised} for comparison. We observe that our descriptors provide both state-of-the-art retrieval results and, depending on descriptor size, also are the fastest for retrieval. Based on these results, dimensionality reduction is recommended, as it both reduces the execution time required for image matching and also decreases the memory footprint. With regards to query expansion, the retrieval execution times should be doubled, since another comparison of the aggregated vector is necessary over the full dataset.

\begin{table}[H]
\caption{Retrieval times (in milliseconds) for different database and descriptor sizes. Values for RFM, SBOW, DN7, DN8 and DBOW are obtained from~\cite{tang2018unsupervised}}.
\label{tab:times}
\centering
\begin{tabular}{crrrrrrrrrr}
\toprule
\begin{tabular}[c]{@{}c@{}}DB size\\ (\# of images)\end{tabular} & \begin{tabular}[c]{@{}c@{}}RFM\\\cite{tang2017sar}\end{tabular} & \begin{tabular}[c]{@{}c@{}}SBOW\\\cite{yang2013geographic}\end{tabular} &
\begin{tabular}[c]{@{}c@{}}Hash\\\cite{demir2016hashing}\end{tabular} &
\begin{tabular}[c]{@{}c@{}}DN7\\\cite{marmanis2016deep}\end{tabular} &
\begin{tabular}[c]{@{}c@{}}DN8\\\cite{marmanis2016deep}\end{tabular} &
\begin{tabular}[c]{@{}c@{}}DBOW\\\cite{tang2018unsupervised}\end{tabular} & \begin{tabular}[c]{@{}c@{}}V-DELF\\ (16K)\end{tabular} &
\begin{tabular}[c]{@{}c@{}}V-DELF\\ (1K)\end{tabular} &
\begin{tabular}[c]{@{}c@{}}V-DELF\\ (512)\end{tabular} &
\begin{tabular}[c]{@{}c@{}}V-DELF\\ (256)\end{tabular} \\
\midrule
50 & 63.10 & 2.20 & 0.92 & 5.80 & 5.70 & 2.30 & 14.09 & 1.07 & 0.97 & 0.61 \\
100 & 81.90 & 6.30 & 2.20 & 17.10 & 17.30 & 6.10 & 54.62 & 3.31 & 3.43 & 1.85 \\
200 & 118.10 & 21.60 & 6.20 & 58.70 & 58.40 & 21.40 & 234.07 & 11.54 & 11.13 & 6.43 \\
300 & 356.60 & 46.00 & 13.30 & 127.40 & 127.80 & 45.90 & 612.52 & 28.18 & 16.56 & 10.72 \\
400 & 396.30 & 79.60 & 19.60 & 223.10 & 224.30 & 79.60 & 1032.22 & 49.01 & 29.72 & 14.87 \\
500 & 440.50 & 124.20 & 29.90 & 346.00 & 344.90 & 123.90 & 1683.54 & 77.83 & 44.90 & 22.98 \\ 
\bottomrule
\end{tabular}
\end{table}

\section{Conclusions}
\label{s5}

In this paper, we have presented a complete RSIR pipeline capable of achieving state-of-the-art results with regards to retrieval precision and computation time. To our best knowledge, this is the first RSIR system to deploy local convolutional features extracted in an end-to-end fashion. We have shown how local convolutional features, together with the VLAD aggregation, yield significantly better performance than existing proposals. This even holds for cases when the features are learned for a different non-RSIR task. We have also evaluated two different attention mechanisms for the purpose of feature selection, and conclude that the multiplicative attention mechanism consistently provides a few percent higher performance. 

Additionally, we introduce low-complexity, yet efficient, query expansion method requiring no user input. This method uses either the \textit{p-sum} or \textit{p-inv} Memory Vector construction, to efficiently aggregate the descriptors of the query image with the descriptors of the top-three candidate images. Expanding the features of the query with those of the best matches increases performance by about 3\%. We have also studied the possibility of compressing our global representations, by reducing the number of visual words in the codebook, or the feature vector size. From these studies we conclude, that even for relatively small descriptor sizes (e.g. 256 or 128), performance is not negatively affected, with some cases presenting even an increased precision. This allows us to couple a high retrieval precision with a low computation time. 

For future work, we want to address the difficulties of our system when dealing with image databases that include fewer classes, or classes with very similar image content (e.g. the ``River'' and ``Harbor'' classes in the SIRI dataset).  More research on other feature extraction architectures or attention mechanisms may yield an even better retrieval performance for such difficult cases. 

\vspace{6pt} 



\authorcontributions{R.I. and C.S. developed the code, performed the experiments and analyzed the results. R.I. wrote the the manuscript. E.B. and P.W. reviewed and edited the manuscript.}

\funding{This research received no external funding.}


\conflictsofinterest{The authors declare no conflict of interest.} 

\reftitle{References}


\externalbibliography{yes}
\bibliography{template}



\end{document}